%% file: sn-article.tex
\documentclass[sn-basic, Numbered]{sn-jnl}

\usepackage{graphicx}
\usepackage{multirow}%
\usepackage{amsmath,amssymb,amsfonts}%
\usepackage{amsthm}%
\usepackage{mathrsfs}%
\usepackage[title]{appendix}%
\usepackage{xcolor}%
\usepackage{textcomp}%
\usepackage{manyfoot}%
\usepackage{booktabs}%
\usepackage{algorithm}%
\usepackage{algorithmicx}%
\usepackage{algpseudocode}%
\usepackage{listings}%
\usepackage[utf8]{inputenc} 
\usepackage[T1]{fontenc}  
\usepackage{scrextend}
\usepackage{tabularx}

\usepackage{hyperref}    
\usepackage{url}      
\usepackage{amsfonts}    
\usepackage{nicefrac}    
\usepackage{microtype}   
\usepackage{lipsum}		
\usepackage{natbib}
\usepackage{doi}
\usepackage{todonotes}
\usepackage{longtable}
\usepackage{rotating}
\usepackage{multirow}
\usepackage{colortbl}
\usepackage[normalem]{ulem}
\useunder{\uline}{\ul}{}
\usepackage{lscape}

\usepackage{lipsum}
\usepackage{pdflscape}
\usepackage{xcolor}
\usepackage[]{acronym}

\usepackage{subcaption}
\usepackage{amssymb}
\usepackage{pifont}
\newcommand{\xmark}{\ding{55}}%

\title[Article Title]{Time Series Analysis in Compressor-Based Machines: A Survey}

\author*[1]{{\fnm{Francesca} \sur{Forbicini} (OrcID: 0009-0007-6971-9872)}}\email{francesca.forbicini@mail.polimi.it}

\author[1]{\fnm{Nicolò Oreste} \sur{Pinciroli Vago} (OrcID: 0000-0001-7906-4987)}\email{nicolooreste.pinciroli@mail.polimi.it}

\author[1]{\fnm{Piero} \sur{Fraternali} (OrcID: 0000-0002-6945-2625)}\email{piero.fraternali@mail.polimi.it}

\affil[1]{\orgdiv{Department of Electronics, Information and Bioengineering}, \orgname{Politecnico di Milano}, \orgaddress{\street{Via Giuseppe Ponzio, 34}, \city{Milan}, \postcode{20133}, \state{Italy}}}

\abstract{
In both industrial and residential contexts, compressor-based machines, such as refrigerators, \ac{HVAC} systems, heat pumps and chillers, are essential to fulfil production and consumers' needs. The diffusion of sensors and IoT connectivity supports the development of monitoring systems that can detect and predict faults, identify behavioural shifts and forecast the operational status of machines and their components. The focus of this paper is to survey the recent research on such tasks as \ac{FD}, \ac{FP}, Forecasting and \ac{CPD} applied to multivariate time series characterizing the operations of compressor-based machines. {These tasks play a critical role in improving the efficiency and longevity of machines by minimizing downtime and maintenance costs and improving the energy efficiency}. Specifically, \ac{FD} detects and diagnoses faults, \ac{FP} predicts such occurrences, forecasting anticipates the future value of characteristic variables of machines and \ac{CPD} identifies significant variations in the behaviour of the appliances, such as a change in the working regime. We identify and classify the approaches to the tasks mentioned above, compare the algorithms employed, highlight the gaps in the current status of the art and discuss the most promising future research directions in the field.}
\begin{document}

\maketitle


\pagebreak
\section*{List of Acronyms}
\input{acronym}

\clearpage
\section{Introduction}
Compressor-based machines are mechanical devices that incorporate compressors as a fundamental component to manipulate gases or fluids, altering their pressure and temperature to perform various functions \cite{Liang2017, Dutta2001}.

{Compressors belong to two main groups: positive displacement compressors and dynamic compressors. Positive displacement compressors in turn comprise reciprocating compressors and rotary compressors \cite{Lu2023}. Dynamic compressors include axial and centrifugal compressors. Positive displacement compressors work by decreasing the volume of a gas in a trapped volume. Dynamic compressors  operate by continuously increasing the momentum of a gas as it flows through them and do not rely on a trapped volume \cite{Hoopes2019}. Positive displacement compressors and dynamic compressors are subject to different failures. The work in \cite{Lu2023} reviews the most common failure mechanisms and patterns for positive displacement compressors:}

\begin{itemize}
    \item {Gas leakage: gas can escape from high-pressure chambers due to empty spaces between machines parts}.
    \item {Overheating: the heat generated during compression affects the temperatures of the working fluid and of the machine components, which may deteriorate}.
    \item {Friction: the repeated contact between moving parts decreases mechanical efficiency and wears the compressor components, such as the gears}.
    \item {Altered vibration: a low operating speed of the suction valve results in valve vibration and periodic fluctuations of the volumetric efficiency}.
\end{itemize}

{The work in \cite{JM4} focuses on dynamic gas compressors, identifies four common issues (friction \cite{Syverud2006}, overheating \cite{Minchev2022}, presence of contaminants \cite{Syverud2007} and corrosion \cite{Sun2015}) and shows that correct lubrication is necessary to mitigate the problems caused by the rotating components.}

In both industrial and residential contexts, compressor-based machines, such as refrigerators \cite{Faria2020, Bansal2021}, air conditioners \cite{Li2022}, heat pumps \cite{Xie2023, Eom2019} and chillers \cite{Fan2020, Yan2017}, are indispensable. Their use is necessary for diverse applications, including food conservation \cite{Holsteijn2018}, medicine \cite{Hearnshaw2007}, air conditioning \cite{Qi2014} and oil and gas treatment \cite{Singhal2014}.
Heat pumps efficiently transfer heat from one location to another in heating and cooling systems \cite{Xu2019_GSHPS, Eom2019}. Chillers extract heat from a liquid through a compression cycle \cite{Fan2020, Yan2017}. \ac{HVAC} systems regulate air quality and temperature in buildings and comprise diverse components \cite{Padmanabh2022, LeCam2017, SalaCardoso2018} whose failure can result in inefficient operations and energy waste. Refrigerators are ubiquitous in households and industrial environments \cite{Bansal2021, Hearnshaw2007} and the capability of anticipating their faults reduces energy consumption and the loss of refrigerated products.

The diffusion of \ac{IoT} data collection \cite{Bansal2021, Floarea2016} has simplified the monitoring of industrial machines, allowing the proactive management of energy consumption and the detection and anticipation of anomalous events \cite{Mourtzis2021}. For example, in a compressor-based machine scenario, data collected with \ac{IoT} sensors can help detect when a refrigerator's door is left open for an extended period \cite{Wu2017}, leading to increased energy consumption, or when the temperature inside the refrigerator deviates from the set range \cite{Perez2019}, indicating a potential issue.

Compressor-based machines provide a significant benchmark for time series analysis techniques due to their pervasive presence in various sectors and the abundance of collected time series data. Such data can be leveraged for various purposes, including predictive maintenance \cite{Kulkarni2018}, energy efficiency optimization and anomaly detection \cite{Zangrando2022}. 
In summary, compressor-based machines offer a rich source of time series data that can be harnessed to enhance their performance, reduce energy wastage and ensure reliable operation, making them a compelling domain for time series analysis and research.

\subsection{Focus of the survey}

This survey is the first one to address the application of \ac{FD}, \ac{FP}, Forecasting and \ac{CPD} on compressor-based machines. \ac{FD} supports predictive maintenance \cite{Nordal2021} because detecting a fault at a given time can help anticipate a more severe occurrence at a successive time, reschedule maintenance and thus reduce unplanned downtime \cite{Zonta2020}. \ac{FP} employs statistical, \ac{ML} and \ac{DL} models to explicitly forecast potential issues before they manifest, allowing for timely interventions \cite{Leukel2022}. Forecasting exploits historical and current data to predict the value of influential status variables \cite{Petropoulos2022, LeCam2017}, which contributes to the efficient operation of machines, ensuring that they function within the optimal parameters' range \cite{Yu2021}. \ac{CPD} focuses on identifying when the system's behaviour undergoes significant changes, which is crucial for determining when a machine transitions from one operational state to another. For instance, it can detect when a refrigerator shifts from an active to an inactive state \cite{van2018detection}.

{The surveyed literature  in \ac{FP}, \ac{FD}, Forecasting and \ac{CPD} highlights several common open issues and the analysis of the existing works allows identifying the most promising research directions.  The main open issues include the lack of annotated datasets containing real data, the absence of public benchmarks for the  evaluation of performances and the scarcity of studies addressing such issues as the generalization and interpretability of the proposed approaches. All the aforementioned issues constitute as many fundamental research directions needed to ensure progress in the field. In addition to such trends,  research could benefit from multi-modal methods able to exploit  diverse data (including not only numerical time series but also images and even sound or smell signals) and from works  that investigate the health status and \ac{RUL} of essential components such as the gears to better characterize  the machine's behaviour.}

\subsection{Methodology}
\label{sec:methodology}

\begin{figure}
 \centering
  \centering
  \includegraphics[width=0.9\textwidth]{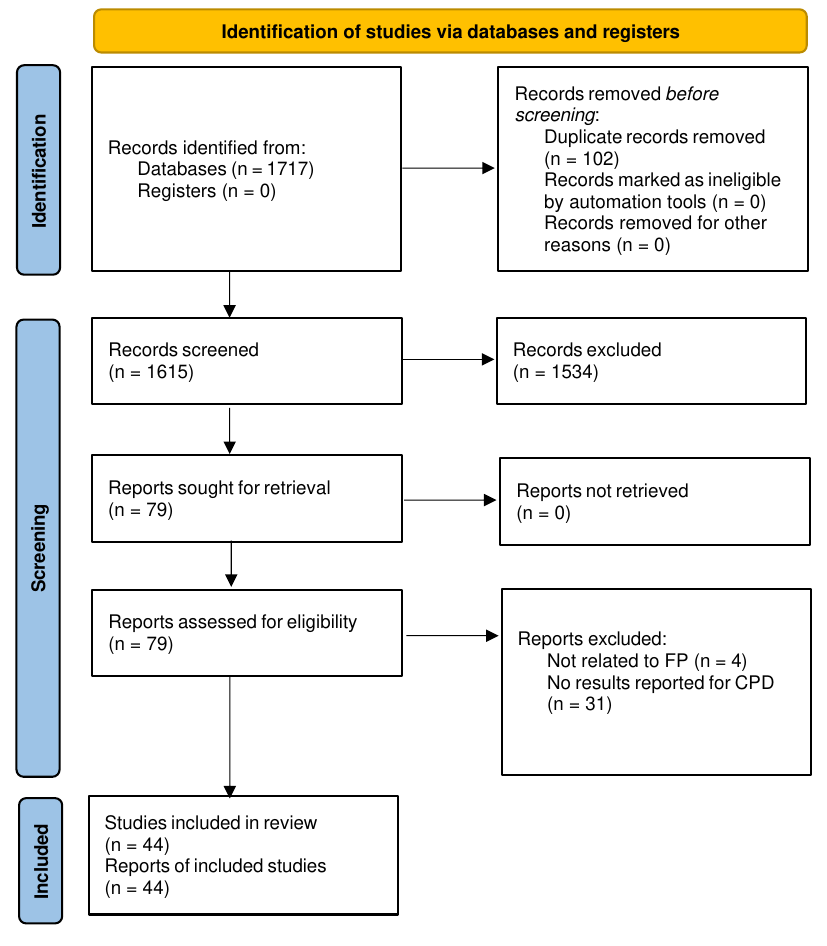}
  \caption{PRISMA flow diagram of the systematic review}
  \label{fig:PRISMA}
\end{figure}

The corpus of the relevant research has been identified by following the PRISMA procedure \cite{Page2021} for systematic reviews. Figure \ref{fig:PRISMA} illustrates the adopted workflow.

\begin{enumerate}
  \item The search was conducted on the Scopus database (for publications between 2016 and 2023) since it has been demonstrated to support bibliographic analysis better than other repositories \cite{Falagas2007}. The search phrases were composed as follows:
  \begin{verbatim}
  <search>:- <task> AND <domain>
  <task>:- fault detection | fault prediction | forecasting 
        | change point detection
  <domain>:- compressor-based machines | chillers | heat pumps | 
        refrigerators | HVAC
  \end{verbatim}
  The search results were filtered to retain only contributions from journals, conferences, and workshops.
  \item The initial corpus, composed of 1717 works, was reduced by removing duplicates; 1615 works were kept. Next, we identified and eliminated the studies based on their relevance by checking each contribution's title, keywords and abstract. The reduced corpus contained 79 contributions.
  \item A final eligibility filter was applied and the full text of the remaining articles was read. In particular, 4 articles were removed because they did not focus on \ac{FP} and 31 articles were removed because they did not report quantitative results for \ac{CPD}. This final step yielded the 44 works considered in this survey.
\end{enumerate}

\begin{figure}
 \centering
 \begin{subfigure}{\textwidth}
  \centering
  \includegraphics[width=0.8\linewidth]{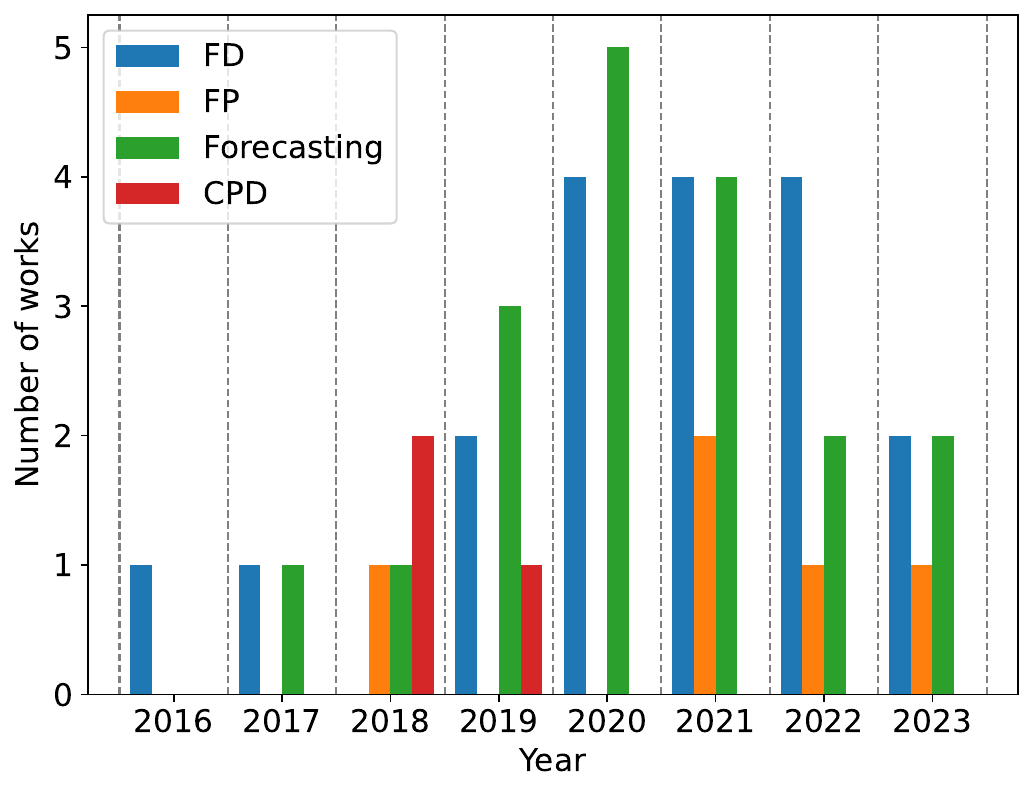}
  \caption{}
  \label{fig:hist_a}
 \end{subfigure}
 \hfill
 \begin{subfigure}{\textwidth}
  \centering
  \includegraphics[width=0.8\linewidth]{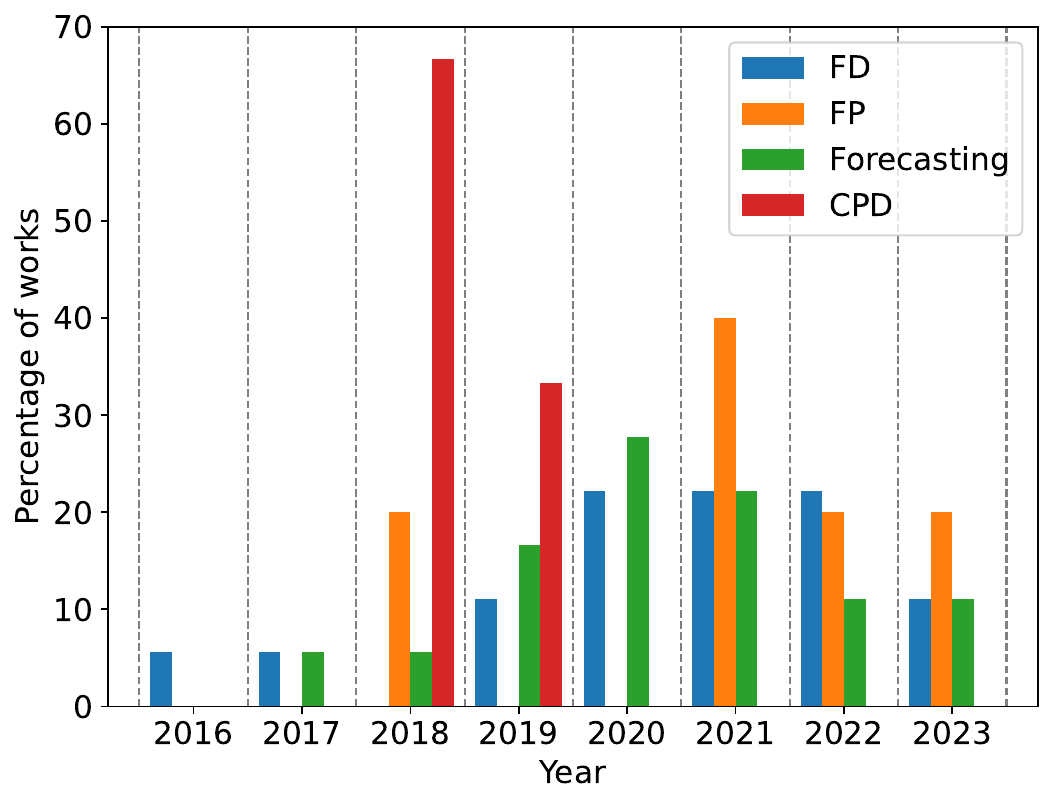}
  \caption{}
  \label{fig:hist_b}
 \end{subfigure}
 \caption{The temporal distribution of works across the 4 analyzed tasks. Fig. 1(a) shows the number of works by year and task. Fig.1(b) shows the percentage of works (with respect to each task) by year and task}
 \label{fig:hist_years}
\end{figure}

Figure \ref{fig:hist_years} shows the distribution of papers through the years, divided by task. Figure \ref{fig:hist_a} shows the number of surveyed works and Figure \ref{fig:hist_b} shows the percentage of surveyed works concerning each task.

\subsection{Contributions}
The contributions of this paper can be summarized as follows:
\begin{itemize}
  \item We survey a total of 44 works addressing the analysis of time data series data of compressor-based machines across the four tasks of \ac{FD}, \ac{FP}, forecasting and \ac{CPD}.
  \item We characterize the existing body of knowledge along the dimensions of application field, machine type, use of real vs simulated data, number of features in the multivariate time series, publication of data with open access, type of supervision and algorithm and type of \ac{GT} annotations. 
  \item We expose the algorithms adopted to treat the four tasks and provide a graphical representation of the pairwise comparisons between algorithms that can be found in the reviewed works (see Figures \ref{fig:fd heatmap}, \ref{fig:fp heatmap}, \ref{fig:forecasting heatmap} and \ref{fig:cpd heatmap}). Given the lack of a comprehensive benchmark to contrast the performances of the various approaches, this representation provides an approximate evaluation of the relative effectiveness of algorithms for the task at hand and the intensity of the efforts for comparing alternative methods. 
  \item We highlight open issues and promising future research directions in \ac{FD}, \ac{FP}, forecasting and \ac{CPD}. Most open issues are methodological. For the analyzed tasks applied to compressor-based machines, there is a scarcity of benchmarks and public datasets for assessing and comparing competing approaches and a lack of agreement on evaluation metrics. The vast body of knowledge and literature in the field demands a more systematic procedure for comparing new approaches with previous works, which is essential for appraising the research progress. The future research directions involve exploring novel approaches for time series analysis. The predictive maintenance challenge for compressor-based machines can benefit from \acp{GNN}, which explicitly take the correlation between signals into account \cite{Xiao2023}, \acp{PINN}, which can combine data-driven and physical models \cite{Borate2023}, and \ac{SR}, which can learn analytically tractable models from data \cite{PhysRevE.94.012214}.
\end{itemize}

The rest of the paper is organized as follows: Section \ref{sec:task} analyzes the characteristics of \ac{FD}, \ac{FP}, forecasting and \ac{CPD} work and the different algorithms used for each task; Section \ref{sec:issues_research_directions} highlights open issues and discusses relevant research directions; and Section \ref{sec:conclusions} draws the conclusions.

\section{Tasks}
\label{sec:task}


\subsection{Fault Detection}
\ac{FD} is the task that identifies faults in a process or system \cite{Park2020}. It can applied to time series acquired in both residential and industrial settings, but often researchers resort to simulated data \cite{Yan2020, Fan2020, Zhu2021} even when studying real systems due to the difficulty of procuring large volumes of data, including real faults \cite{Abhiraman2023}. An approach based on synthetic time series captures the complexity of real-world data only partially, leading to a gap between research findings and practical applications \cite{Nie2023, Yan2022, Soltani2022}. 
\ac{FD} for industrial compressor-based machines poses specific challenges due to the configuration of appliances based on the requirements of the specific production process, which leads to different behaviours \cite{Soltani2020}. For this reason, \ac{FD} techniques must exhibit robustness to configuration changes and adaptation to different operating conditions.

\subsection{Fault Prediction}

\ac{FP} is the process of analyzing historical data to predict future faults in a system \cite{Zhang2018}. This task involves finding deviations from normal behaviour and identifying the ones leading to future faults. 
As \ac{FD} also \ac{FP} is challenged by the difficulty of procuring sufficiently large datasets incorporating real faults. 

A specific aspect of \ac{FP} is that accuracy is affected by the temporal horizon of the predictions and the amount of past data considered to build them. The work in \cite{Leukel2022} systematically studies such dependency by varying both the amount of historical data, or \ac{RW} and the prediction horizon, or \ac{PW}. In general, the choice of the \ac{RW} and \ac{PW} is challenging, as they depend on the specific case study \cite{Bonnevay2019, Colone2019, FigueroaBarraza2020, Khorsheed2020, Kusiak2012, Li2014, Leukel2022, Kaparthi2020, Leahy2018, Proto2019}. A sufficiently large \ac{PW} size is crucial for implementing predictive maintenance effectively by enabling timely interventions. However, an extended \ac{PW} can detriment the ability to make accurate predictions and increase the rate of false positives. 
Various factors, including sensor noise, transient conditions, or non-critical deviations, can cause false alarms \cite{Bansal2021}. Distinguishing between false alarms and precursors to failures remains a significant research issue because false alarms can produce unnecessary maintenance and operational disruptions \cite{Jiang2023, Padmanabh2021}. In compressor-based machines, a gradual degradation over time of specific components that affects operational performance can be observed before an actual failure materializes \cite{Jiang2023}. Such degradation can reduce machine performance well before an actual fault occurs and identifying such early signs of deterioration is challenging \cite{Padmanabh2022, Jiang2023, Padmanabh2021} and requires proper \ac{CPD} methods.

\subsection{Forecasting}

Forecasting is the task that exploits past observations to develop a model for predicting the future value of variables \cite{Zhang2003}. As \ac{FP}, forecasting requires the definition of a prediction horizon, which impacts the strength of the correlation between the observed data and the predicted data \cite{Kim2020, Liu2023, Xu2021}. The works in \cite{Xie2020, Petroanu2020} show that accuracy is higher when predicting short-term outcomes and decays with the extension into the future of the prediction horizon. Considering multivariate time series data, it is challenging to capture correlations between different features to predict the behaviour of target variables and ad-hoc methods have been developed \cite{Ahmad2019, Li2022, LeCam2017}. 
In this context, the accurate forecast of the load profiles, such as the heating and cooling load of \ac{HVAC} systems or refrigerators, is crucial to optimize energy consumption \cite{Ahmad2019, Xie2020} and improve the operational efficiency \cite{Li2022, Xie2023}

\subsection{Change Point Detection}

\ac{CPD} is the problem of finding time points at which the properties of a time series change, which is a clue that the underlying process that generates the data may have changed too \cite{doi:10.1137/1.9781611972795.34}. As in the other analyzed tasks, \ac{CPD} benefits from labelled data for training and evaluation \cite{Touzani2019, Pereira2018}, but such data are scarce. Therefore, some works resort to unlabelled data and apply unsupervised learning techniques \cite{van2018detection}, which require the definition of a suitable threshold to determine when a behaviour change has occurred \cite{van2018detection, Pereira2018}. The choice of the threshold value can significantly impact performances and may require domain knowledge \cite{van2018detection}. In compressor-based machines, the primary focus is detecting changes in the machine's activity, such as on-off switching \cite{van2018detection, Pereira2018}. The change points might correspond to different modes or states of the machine's operation, characterized by variations in power consumption, temperature, or other relevant parameters \cite{Touzani2019, van2018detection, Pereira2018}. Also in this specific case, the lack of labels denoting status changes \cite{van2018detection, Pereira2018} imply the use of unsupervised techniques such as clustering \cite{van2018detection, Touzani2019}, which in turn require the selection of an adequate threshold.

{\subsection{Boundaries between different tasks}}
{\ac{FD}, \ac{FP}, Forecasting and \ac{CPD} have distinct boundaries and specific objectives. However, their practical applications often overlap. Most of the surveyed papers clearly identify the task they address.}

\begin{itemize}
    \item {\ac{FD} vs \ac{CPD}: some works use \ac{CPD} to detect faults \cite{Kawahara2007, DAngelo2011, KJ2021}. However, the two tasks have different goals: \ac{CPD} intends to identify shifts and changes in the monitored behaviour, whereas \ac{FD} aims to detect faults. Behavioural changes are not necessarily a warning signal and can be observed in diverse situations \cite{Suhaila2017, Tsirigos2005}. Moreover, faults can be abrupt \cite{HanQiuZhang2001} and not preceded by early warning signals, making \ac{CPD} approaches that identify gradual changes ineffective.}

    \item {Forecasting vs \ac{FP}: \ac{FP} predicts forthcoming  failures based on historical data.  Forecasting predicts future values of various operational characteristics of a machine and is not limited to faults. For this reason, some works use forecasting as an intermediate step to predict faults \cite{SopelsaNeto2022, Baptista2018, deljac2011comparison}. The work in \cite{Zhengdao2008} defines \ac{FP} as the combination of \ac{FD} and forecasting and the work in \cite{Li2021} states that \ac{FP} aims at building a classification model for different fault types. In summary, \ac{FP} has a more specific objective than Forecasting and this survey categorizes as \ac{FP} all works that investigate approaches usable to predict faults.}

    \item {\ac{FD} vs \ac{FP}: \ac{FD} identifies faults when they occur whereas  \ac{FP} predicts faults before they occur. In general, the tasks of detecting a fault and predicting a future faulty behaviour are distinct, even if related \cite{Yao2023, Glvez2021}. \ac{FD} can be used to find early signals suggesting the insurgence of more severe faults, but it is not designed to provide information on whether and when they will happen. For instance, the work in \cite{Bangalore2015} shows that a severe fault that required the replacement of a mechanical component was preceded by the spalling in a bearing. The survey in \cite{Leukel2021}, instead, focuses on \ac{FP} and shows that most \ac{FP} works consider well-defined \acp{PW} (i.e., future time horizons).}

\end{itemize}

\subsection{Datasets}
\label{sec:Datasets}

The progress of research in data-driven predictive algorithms obviously depends on the availability of quality data to train, evaluate and compare approaches. In other domains where data-driven methods are employed, such as natural language and image processing, the research community has produced over time a vast body of datasets and benchmarks, which have played a fundamental role in the progress of the state-of-the-art.
This is less the case in the realm of time series analysis in general and of industrial and compressor-based machines in particular. Most surveyed works employ a private dataset, which is neither published nor used for comparison with other approaches. For example, the work in \cite{Nie2023} develops the simulation model of an air conditioner using the Amesim software\footnote{\url{https://plm.sw.siemens.com/it-IT/simcenter/systems-simulation/amesim/}}, but the dataset is not public.
As a notable exception, a few 
 works \cite{Yan2020, Yan2017, Fan2020, Lee2022, Nie2023, Li2016, Yan2022} adopt the \acs{ASHRAE} 1043-RP project dataset \cite{comstock1999experimental}, which simulates the behaviour of 90-ton centrifugal chillers with 64 parameters. This dataset contains seven types of faults and different \acp{SL}. Table \ref{tab:faults_ashrae} summarizes the fault names, the quantity whose change leads to faults (the variable feature) and the associated \acp{SL}. Each percentage refers to the variation of the variable feature necessary to obtain a fault with the \ac{SL} indicated in the corresponding column.

\begin{table}[h]
\centering
\caption{\ac{ASHRAE} 1043-RP seven faults, the variable feature causing them and the variation of the variable feature needed for reaching different \acp{SL}}
\label{tab:faults_ashrae}
\begin{tabular}{p{0.2\textwidth}cccccc}
\toprule
\textbf{Fault name} & \textbf{Variable feature} & \textbf{\ac{SL}1} & \textbf{\ac{SL}2} & \textbf{\ac{SL}3} & \textbf{\ac{SL}4} \\ \midrule
Flow water of condenser Insufficient (FWC) & Condenser water flow rate &-10\% & -20\% & -30\% & -40\% \\ \midrule
Flow water of evaporator Insufficient (FWE) & Evaporator water flow rate&-10\% & -20\% & -30\% & -40\% \\ \midrule
Refrigerant leak (RL) & Refrigerant Charge &-10\% & -20\% & -30\% & -40\% \\ \midrule
Refrigerant over (RO)& Refrigerant Charge & +10\% & +20\% & +30\% & +40\% \\ \midrule
Condenser fouling (CF) & Plugged tubes in the condenser & -12\% & -20\% & -30\% & -45\% \\ \midrule
Excessive over (EO) & Oil charge & +14\% & +32\% & +50\% & +68\% \\ \midrule
Non-condensable gas contained (NC) & Adding more nitrogen to the refrigerant &+1\% & +2\% & +3\% & +5\% \\ \bottomrule
\end{tabular}

\end{table}

\subsection{Approaches and their characterization}
\label{sec:Tasks and methods}

Sections \ref{sec:app_fd}, \ref{sec:app_fp}, \ref{sec:app_forecasting} and \ref{sec:app_cpd} illustrate the research works on \ac{FD}, \ac{FP}, Forecasting and \ac{CPD}, respectively. For each study, datasets and algorithms are classified based on the following dimensions:

\begin{itemize}
  \item Real Data: Indicates whether the data are real (Y) or simulated (N)
  \item Field: Specifies if the system presented in the study belongs to an industrial, domestic or different environment.
  \item Machine Type: Indicates the type of machine considered (e.g., Chiller, Heat pump, \ac{HVAC} and Refrigerator).
  \item Public Dataset: Specifies whether the dataset is publicly available (Y) or private (N).
  \item Features: Indicates the number of variables in the dataset before any preprocessing in the research. If the exact number of features is not reported, it is denoted as \textbf{M}. 
  \item Supervision: Specifies the presence of supervision:
    \begin{itemize}
    \item \textbf{S (Supervised):} The research exploits supervised learning.
    \item \textbf{U (Unsupervised):} The research involves unsupervised learning only.
  \end{itemize}
  \item \ac{GT}: Explains how the ground truth in a dataset has been obtained:
  \begin{itemize}
    \item \textbf{S (Simulated):} Faults are simulated and the \ac{GT} is produced by the simulation.
    \item \textbf{R (Rule-based):} \ac{GT} is determined based on rules (e.g., a fault if the temperature exceeds a threshold).
    \item \textbf{E (Experts):} \ac{GT} is labeled by experts in the field.
    \item \textbf{I (Included):} In forecasting tasks, where the time series is split into historical and predicted data, the \ac{GT} is the portion of the time series to be predicted.
  \end{itemize}
  \item Algorithm: Name of the top-performing algorithms illustrated in the paper.
\end{itemize}

For each task, the algorithms presented in the surveyed works are identified. Their relative performances are evaluated approximately by providing a graphical representation summarizing the pairwise comparisons performed in the reviewed papers. Algorithms belonging to the same family are grouped (e.g., \ac{SVM} indicates the family of \ac{SVM}-based algorithms, possibly with ad-hoc preprocessing).

\subsubsection{Fault Detection}
\label{sec:app_fd}

\ac{FD} methods cover a wide range of techniques rooted in both \ac{ML} and \ac{DL} algorithms. Among the commonly employed \ac{ML} algorithms are \ac{SVM} \cite{Soltani2022, Yan2021, Nie2023, Yan2020, Fan2020}, various \ac{SVM} variants \cite{Yan2017, Dey2020}, XGBoost \cite{Bansal2021} and \ac{LDA} \cite{Soltani2022, Li2016}. A methodological contribution is presented in \cite{Lee2022}, which makes a comparative analysis of different \ac{ML} techniques. The advent of \ac{DL} can also be appreciated in the \ac{FD} works, which employ algorithms such \ac{1D-CNN} \cite{Kuendee2019, Yan2022}, \ac{LSTM} \cite{Tian2021}, Siamese Networks \cite{Chen2023}, \acp{CNN} \cite{Abhiraman2023, Soltani2020, Eom2019} and \acp{DRNN} \cite{Taheri2021}. The work in \cite{Zhu2021} harnesses the power of \ac{DL} and \ac{TL}, introducing the innovative \ac{DANN} for \ac{FD}.

Table \ref{tab:fault_detection_results} summarizes the datasets, algorithms and metrics used by the \ac{FD} works. Note that for \cite{Yan2020}, which uses different \ac{SVM}-based algorithms on a varying number of features, the table reports \ac{SVM} as the employed family of algorithms. In the case of \cite{Lee2022}, the table classifies the data as not real because the authors use in part the \ac{ASHRAE} dataset for both training and evaluation. When some works, such as \cite{Nie2023, Abhiraman2023, Taheri2021, Soltani2022}, employ multiple datasets, the table includes a distinct row for each dataset.

\begin{sidewaystable}
 \caption{An overview of \ac{FD} works and their characteristics. All approaches are supervised}
\label{tab:fault_detection_results}
\centering
\begin{tabular*}{\textheight}{@{\extracolsep\fill}ccccccccc}

\\ \toprule   
\multicolumn{1}{c}{\textbf{Paper}} & \multicolumn{1}{c}{\textbf{Year}} & \multicolumn{1}{c}{\textbf{Real data}} & \multicolumn{1}{c}{\textbf{Field}} & \multicolumn{1}{c}{\textbf{Machine type}} & \multicolumn{1}{c}{\textbf{Public dataset}} & \multicolumn{1}{c}{\textbf{Features}} & \multicolumn{1}{c}{\textbf{\ac{GT}}} & \multicolumn{1}{c}{\textbf{Algorithm}} \\ \midrule

\multicolumn{1}{c}{{\cite{Li2016}}} & \multicolumn{1}{c}{2016} & \multicolumn{1}{c}{N} & \multicolumn{1}{c}{Industrial} & \multicolumn{1}{c}{Chiller} & \multicolumn{1}{c}{Y\footnote{\label{ashrae_dataset}\url{https://www.worldcat.org/it/title/experimental-data-from-fault-detection-and-diagnostic-studies-on-a-centrifugal-chiller/oclc/738032681}}
} & \multicolumn{1}{c}{64} & \multicolumn{1}{c}{S} & \multicolumn{1}{c}{{\color[HTML]{000000} \acs{LDA}}} \\ \midrule
\multicolumn{1}{c}{{\cite{Yan2017}}} & \multicolumn{1}{c}{2017} & \multicolumn{1}{c}{N} & \multicolumn{1}{c}{Industrial} & \multicolumn{1}{c}{Chiller} & \multicolumn{1}{c}{Y\footref{ashrae_dataset}} & \multicolumn{1}{c}{64} & \multicolumn{1}{c}{S} & \multicolumn{1}{c}{{\color[HTML]{000000} \acs{EKF}-\acs{ROSVM}}} \\
\midrule
\multicolumn{1}{c}{{\cite{Eom2019}}} & \multicolumn{1}{c}{2019} & \multicolumn{1}{c}{N} & \multicolumn{1}{c}{Industrial} & \multicolumn{1}{c}{Heat pump} & \multicolumn{1}{c}{N} & \multicolumn{1}{c}{28} & \multicolumn{1}{c}{S} & \multicolumn{1}{c}{{\color[HTML]{000000} \acs{CNN}-ReLU-He normal}} \\ \midrule

\multicolumn{1}{c}{\cite{Kuendee2019}} & \multicolumn{1}{c}{2019} & \multicolumn{1}{c}{Y} & \multicolumn{1}{c}{Industrial} & \multicolumn{1}{c}{Refrigerator} & \multicolumn{1}{c}{N} & \multicolumn{1}{c}{6} & \multicolumn{1}{c}{E} & \multicolumn{1}{c}{{\color[HTML]{000000} \acs{CNN}}} \\ \midrule
\multicolumn{1}{c}{{\cite{Yan2020}}} & \multicolumn{1}{c}{2020} & \multicolumn{1}{c}{N} & \multicolumn{1}{c}{Industrial} & \multicolumn{1}{c}{Chiller} & \multicolumn{1}{c}{Y\footref{ashrae_dataset}} & \multicolumn{1}{c}{64} & \multicolumn{1}{c}{S} & \multicolumn{1}{c}{{\color[HTML]{000000} \acs{CWGAN}-\acs{SVM}}} \\
\midrule

\multicolumn{1}{c}{{\cite{Fan2020}}} & \multicolumn{1}{c}{2020} & \multicolumn{1}{c}{N} & \multicolumn{1}{c}{Industrial} & \multicolumn{1}{c}{Chiller} & \multicolumn{1}{c}{Y\footref{ashrae_dataset}} & \multicolumn{1}{c}{64} & \multicolumn{1}{c}{S} & \multicolumn{1}{c}{{\color[HTML]{000000} \acp{SVM}}} \\ 
\midrule
\multicolumn{1}{c}{{\cite{Soltani2020}}} & \multicolumn{1}{c}{2020} & \multicolumn{1}{c}{Y} & \multicolumn{1}{c}{Industrial} & \multicolumn{1}{c}{Refrigerator} & \multicolumn{1}{c}{N} & \multicolumn{1}{c}{14} & \multicolumn{1}{c}{S} & \multicolumn{1}{c}{{\color[HTML]{000000} \acs{CNN}}} \\
 & & \multicolumn{1}{c}{Y} & \multicolumn{1}{c}{Industrial} & \multicolumn{1}{c}{Refrigerator} & \multicolumn{1}{c}{N} & \multicolumn{1}{c}{14} & \multicolumn{1}{c}{S} & \multicolumn{1}{c}{{\color[HTML]{000000} \acs{CNN}}} \\ \midrule
 \multicolumn{1}{c}{{\cite{Dey2020}}} & \multicolumn{1}{c}{2020} & \multicolumn{1}{c}{Y} & \multicolumn{1}{c}{Building} & \multicolumn{1}{c}{\ac{HVAC}} & \multicolumn{1}{c}{N} & \multicolumn{1}{c}{6} & \multicolumn{1}{c}{R} & \multicolumn{1}{c}{{\color[HTML]{000000} \acs{MC-SVM}}} \\ \midrule
\multicolumn{1}{c}{{\cite{Zhu2021}}} & \multicolumn{1}{c}{2021} & \multicolumn{1}{c}{N} & \multicolumn{1}{c}{Industrial} & \multicolumn{1}{c}{Chiller} & \multicolumn{1}{c}{N} & \multicolumn{1}{c}{16} & \multicolumn{1}{c}{S} & \multicolumn{1}{c}{{\color[HTML]{000000} \acs{DANN}}} \\ 
\midrule
\multicolumn{1}{c}{{\cite{Taheri2021}}} & \multicolumn{1}{c}{2021} & \multicolumn{1}{c}{N} & \multicolumn{1}{c}{Building} & \multicolumn{1}{c}{\ac{HVAC}} & \multicolumn{1}{c}{Y\footnote{\label{SZ/MZ-AHU}\url{https://flexlab.lbl.gov/}}} & \multicolumn{1}{c}{18} & \multicolumn{1}{c}{S} & \multicolumn{1}{c}{{\color[HTML]{000000} \acs{S,DTO-DRNN}}} \\ 
 & & \multicolumn{1}{c}{N} & \multicolumn{1}{c}{Building} & \multicolumn{1}{c}{\ac{HVAC}} & \multicolumn{1}{c}{Y\footref{SZ/MZ-AHU}} & \multicolumn{1}{c}{18} & \multicolumn{1}{c}{S} & \multicolumn{1}{c}{{\color[HTML]{000000} \acs{S,DTO-DRNN}}}\\ \midrule
\multicolumn{1}{c}{{\cite{Tian2021}}} & \multicolumn{1}{c}{2021} & \multicolumn{1}{c}{N} & \multicolumn{1}{c}{Industrial} & \multicolumn{1}{c}{Refrigerator} & \multicolumn{1}{c}{N} & \multicolumn{1}{c}{13} & \multicolumn{1}{c}{S} & \multicolumn{1}{c}{{\color[HTML]{000000} \acs{LSTM}}} \\ \midrule
\multicolumn{1}{c}{{\cite{Yan2021}}} & \multicolumn{1}{c}{2021} & \multicolumn{1}{c}{N} & \multicolumn{1}{c}{Industrial} & \multicolumn{1}{c}{Chiller} & \multicolumn{1}{c}{Y\footref{ashrae_dataset}} & \multicolumn{1}{c}{64} & \multicolumn{1}{c}{S} & \multicolumn{1}{c}{{\color[HTML]{000000} \acs{CWGAN}-GANomaly-\acs{SVM}}} \\ \midrule
\multicolumn{1}{c}{\cite{Soltani2022}} & \multicolumn{1}{c}{2022} & \multicolumn{1}{c}{N} & \multicolumn{1}{c}{Industrial} & \multicolumn{1}{c}{Refrigerator} & \multicolumn{1}{c}{N} & \multicolumn{1}{c}{14} & \multicolumn{1}{c}{S} & \multicolumn{1}{c}{{\color[HTML]{000000} \acs{SVM}}} \\
& & \multicolumn{1}{c}{N} & \multicolumn{1}{c}{Industrial} & \multicolumn{1}{c}{Refrigerator} & \multicolumn{1}{c}{N} & \multicolumn{1}{c}{14} & \multicolumn{1}{c}{S} & \multicolumn{1}{c}{{\color[HTML]{000000} \acs{SVM}}} \\
& & \multicolumn{1}{c}{N} & \multicolumn{1}{c}{Industrial} & \multicolumn{1}{c}{Refrigerator} & \multicolumn{1}{c}{N} & \multicolumn{1}{c}{14} & \multicolumn{1}{c}{S} & \multicolumn{1}{c}{{\color[HTML]{000000} \acs{LDA}-\acs{SVM}}}\\ \midrule
\multicolumn{1}{c}{{\cite{Lee2022}}} & \multicolumn{1}{c}{2022} & \multicolumn{1}{c}{N, N} & \multicolumn{1}{c}{Industrial} & \multicolumn{1}{c}{Chiller} & \multicolumn{1}{c}{Y\footref{ashrae_dataset},N} & \multicolumn{1}{c}{64, M} & \multicolumn{1}{c}{S, E} & \multicolumn{1}{c}{{\color[HTML]{000000} \acs{DJ}}}\\ 
 & & & & & & & & \multicolumn{1}{c}{{\color[HTML]{000000} \acs{ANN}}} \\ \midrule

 \multicolumn{1}{c}{{\cite{Yan2022}}} & \multicolumn{1}{c}{2022} & \multicolumn{1}{c}{N} & \multicolumn{1}{c}{Industrial} & \multicolumn{1}{c}{Chiller} & \multicolumn{1}{c}{Y\footref{ashrae_dataset}} & \multicolumn{1}{c}{64} & \multicolumn{1}{c}{S} & \multicolumn{1}{c}{{\color[HTML]{000000} \acs{1D-CNN}}} \\ \midrule
\multicolumn{1}{c}{{\cite{Abhiraman2023}}} & \multicolumn{1}{c}{2022} & \multicolumn{1}{c}{N} & \multicolumn{1}{c}{Industrial} & \multicolumn{1}{c}{Refrigerator} & \multicolumn{1}{c}{N} & \multicolumn{1}{c}{6} & \multicolumn{1}{c}{S} & \multicolumn{1}{c}{{\color[HTML]{000000} \acs{CNN}}} \\
& & \multicolumn{1}{c}{N} & \multicolumn{1}{c}{Industrial} & \multicolumn{1}{c}{Refrigerator} & \multicolumn{1}{c}{N} & \multicolumn{1}{c}{6} & \multicolumn{1}{c}{S} & \multicolumn{1}{c}{{\color[HTML]{000000} \acs{CNN}}} \\\midrule

\multicolumn{1}{c}{{\cite{Nie2023}}} & \multicolumn{1}{c}{2023} & \multicolumn{1}{c}{N} & \multicolumn{1}{c}{Industrial} & \multicolumn{1}{c}{Chiller} & \multicolumn{1}{c}{Y\footref{ashrae_dataset}} & \multicolumn{1}{c}{64} & \multicolumn{1}{c}{S} & \multicolumn{1}{c}{{\color[HTML]{000000} \acs{SVM}-ReliefF-\acs{RFECV}}} \\ 
 & & \multicolumn{1}{c}{N} & \multicolumn{1}{c}{Industrial} & \multicolumn{1}{c}{Chiller} & \multicolumn{1}{c}{N} & \multicolumn{1}{c}{24} & \multicolumn{1}{c}{S} & \multicolumn{1}{c}{{\color[HTML]{000000} \acs{SVM}-ReliefF-\acs{RFECV}}} \\ \midrule

\multicolumn{1}{c}{{\cite{Chen2023}}} & \multicolumn{1}{c}{2023} & \multicolumn{1}{c}{N} & \multicolumn{1}{c}{Industrial} & \multicolumn{1}{c}{Chiller} & \multicolumn{1}{c}{Y\footnote{\url{https://www.techstreet.com/standards/rp-1312-tools-for-evaluating-fault-detection-and-diagnostic-methods-for-air-handling-units?product_id=1833299}}} & \multicolumn{1}{c}{15} & \multicolumn{1}{c}{S} & \multicolumn{1}{c}{{\color[HTML]{000000} \acs{SNN}-\acs{LSTM}}} \\ 
\bottomrule
\end{tabular*}

\end{sidewaystable}

Most works are applied to industrial machines ($\approx 87\%$), which underlines the importance of \ac{FD} in the industrial sector. Only a minority of the works ($\approx 42\%$) use a public dataset and most of them exploit the same one (\ac{ASHRAE}). Simulated data are prevalent (in $\approx 83\%$ of the cases) and no work considers public datasets made of real time series, which would be beneficial for better evaluating the effectiveness of different algorithms in practical conditions. All works analyze multivariate time series, underscoring the importance of collecting several variables in complex physical systems. 

In \cite{Soltani2022} \ac{FD} is performed on refrigerator systems data with \ac{CNN}, \ac{LDA} and \ac{SVM} classifiers. The algorithms are tested both on the original time series and on a dimensionally reduced version created with \ac{LDA} and \ac{PCA}. Performance evaluation highlights \ac{SVM} (99.6\% accuracy) and \ac{LDA} used as a classification method (99.8\% accuracy) as superior, with the \ac{SVM} 
and \ac{CNN} classifiers applied to the reduced time series showing varied results. \ac{PCA}-\ac{SVM} suffers from class separation issues due to the projection onto a lower-dimensional space and has difficulties in distinguishing certain types of faults in different operating conditions (e.g. different ambient temperature and heat load). \ac{LDA}-\ac{SVM} reduces the training time with respect to other approaches but achieves a lower accuracy ($96.6\%$) than \ac{LDA}. The work in \cite{Li2016} employs a two-stage \ac{LDA} to detect the seven faults in the \ac{ASHRAE} 1043-RP dataset. First, it clusters training data in one normal cluster and seven fault clusters (one per fault type). Then, \ac{LDA} reduces data dimensionality. Manhattan distance is computed on \ac{LDA} output between each point and the corresponding cluster centre to evaluate whether sequences of the original data points are (1) close to a fault cluster centre and (2) within the limits of that cluster (i.e., they correspond to faulty sequences). Finally, data are clustered based on \acp{SL} for each fault type using a similar procedure.

The work in \cite{Fan2020} employs data taken by the \ac{ASHRAE} 1043-RP dataset to detect faults using a varying amount of input variables. The dataset contains 64 variables, 48 of which are measured by sensors. The work in \cite{Wang2018} shows that only 8 sensors are essential in large chillers to detect faults (namely, five temperature sensors, two pressure sensors and the compressor input power). A first experiment trains \ac{SVM}-3, an \ac{SVM} with only 3 sensors as input (Compressor input power, Evaporating pressure and Condensing pressure), regarded as the most important for detecting chillers faults. The second experiment employs 8 sensors to train \ac{SVM}-8. The last experiment uses all the 64 variables to train \ac{SVM}-64. The three \ac{SVM} variants are all evaluated using three datasets with different numbers of features, 8,000 train samples and 4,000 test samples. \ac{SVM}-3 achieves an overall accuracy of 88.34\%, \ac{SVM}-8 of 96.68\% and \ac{SVM}-64 of 99.36\%. Increasing the number of sensors (\ac{SVM}-64) leads to marginal accuracy improvements. 

The \ac{ASHRAE} 1043-RP dataset is also used in \cite{Yan2017}, which (1) selects 11 out of 64 features using the ReliefF method and an adaptive \ac{GA}, (2) employs \ac{EKF} to denoise and stationarize the time series and (3) trains \ac{ROSVM}, an \ac{SVM} variant. \ac{ROSVM} differs from the standard \ac{SVM} because (1) it learns to characterize the normal region in the data space, classifying the points outside it as anomalous and (2) during training, new samples are continuously taken in input into \ac{OSVM} to refine the normal region bounds. ReliefF selects the features with maximal information. Then, the selected features are further reduced by an adaptive \ac{GA}. Next, the statistical method \ac{EKF} \cite{mulumba2014kalman} removes the noise from the time series and makes them stationary. \ac{ROSVM} is trained on the output of \ac{EKF} to detect all seven faults at different \acp{SL}. The preprocessing steps yield higher accuracy than the application of \ac{ROSVM} on the original data, highlighting the importance of feature selection and filtering.

The work in \cite{Yan2021} addresses the class-imbalance problem of the \ac{ASHRAE} 1043-RP dataset by generating synthetic faults. First, \ac{CWGAN} generates a large number of faults with different \acp{SL} by using a small amount of real \ac{ASHRAE} data. Then, it compares three fault selection algorithms (the ensemble algorithm proposed in \cite{zhong2019energy}, GANomaly and \ac{VAE}) to select a subset of such faults. The ensemble algorithm is a \ac{GAN} that evaluates the Wasserstein distance between the generated and original sample space. GANomaly is an anomaly detection algorithm that comprises two encoders and a decoder. It computes the potential encoder error between the original and reconstructed samples to discard artificial samples. \ac{VAE} computes the reconstruction error between the original and reconstructed samples and discards artificial samples. For each fault selection algorithm, the selected faults are added to the dataset with different percentages (10\%, 30\% and 50\%). Six \ac{FD} algorithms are trained (\ac{SVM}, \ac{RF}, \ac{DT}, \ac{NB}, \ac{MLP} and \ac{KNN}) to detect the generated faults. \ac{SVM} achieves higher accuracy and $F_1$ score compared with the other \ac{FD} algorithms.

Another work addressing the imbalance issue in the \ac{ASHRAE} dataset is \cite{Yan2020}. First, \ac{CWGAN} generates synthetic faults with different \acp{SL}. Then, \ac{SVM} is compared to \ac{RF}, \ac{DT}, \ac{BN}, \ac{KNN} and \ac{LR} on binary and multi-class classification in the generated dataset. In both cases, \ac{SVM} has greater accuracy than the other classifiers. Introducing synthetic training samples using \ac{CWGAN} yields higher accuracy than that obtained with the original dataset. 

The work in \cite{Nie2023} uses the \ac{ASHRAE} 1043-RP dataset and the Amesim dataset, a time series collection with 24 features from a simulated air conditioning system containing three types of faults. First, a subset of features is selected using the ReliefF and \ac{RFECV} algorithms. ReliefF computes the weights for each feature. Those features and their weights are fed as input to \ac{RFECV}, which (1) ranks the features according to their weights and (2) selects the combination of features leading to the highest classification accuracy. Classification accuracy is evaluated using the \ac{RF} algorithm and 10 folds. The feature subset leading to the highest accuracy is selected as the best. \ac{RFECV} is beneficial because ReliefF cannot remove redundant features, as it can only capture the strong correlation between them. On the other hand, \ac{RFECV} is time-consuming because it performs cross-validation to select the optimal features subset. The described feature selection approach reduces \ac{ASHRAE} features from 64 to 24 and those in Amesim from 24 to 6. After feature selection, three \ac{ML} classifiers (\ac{SVM}, \ac{RF}, \ac{KNN}) are compared to classify the seven faults with their different \acp{SL} in the \ac{ASHRAE} dataset and the three faults in the Amesim dataset. \ac{SVM} shows higher accuracy than \ac{RF} and \ac{KNN} in both datasets.

The work in \cite{Dey2020} uses a two-step approach to detect six faults in a dataset acquired from \ac{HVAC} units composed of a heating and cooling coil and a fan. Data are clustered to distinguish the normal and faulty behaviours. The generated clusters are given in input to the \ac{MC-SVM} classifier, which is compared to \ac{KNN} with $k=1$ (1NN) and $k=3$ (3NN) by considering the same test set, different proportions of the train set (10\%, 20\% and 30\%) and different aggregations of the events (daily, weekly, monthly and randomly averaged).  \ac{MC-SVM} achieves slightly better precision and recall than 1NN and 3NN.

The works in \cite{Zhu2021, Lee2022} use \ac{TL} to adapt models trained on information-rich datasets to new datasets with limited data. 
The work in \cite{Lee2022} detects six types of faults (e.g., refrigerant leakage and pipeline restriction) in a dataset acquired from 100 refrigeration systems. First, the \ac{ASHRAE} 1043-RP is processed using auto-scale \ac{TL}. Then, the processed dataset is used for training five fault identification algorithms (\ac{DT}, \ac{DJ}, \ac{LR}, \acp{ANN} and \ac{OVA}) for different types of faults and fault frequencies. The combination of \ac{TL} and the fault detection algorithm leads to high accuracy results (higher than 95\% for all fault types but refrigerant leakage, which reaches 83.3\%), surpassing state-of-the-art accuracy.
The work in \cite{Zhu2021} uses a simulated time series with four faults and employs a \ac{DANN} to detect failures in a target chiller using data from a source chiller. First, data are standardized to make them comparable for the source and target chillers. Then, the same feature selection is applied to both chillers. Finally, the \ac{DANN} is trained on the standardized normal and fault data of the source chiller and the standardized normal data of the target chiller. The proposed method shows greater accuracy than \ac{KNN}, \ac{SVM}, \ac{DT}, \ac{RF} and \ac{BN} thanks to its ability to use prior knowledge.

The works in \cite{Kuendee2019, Yan2022} apply a \ac{1D-CNN} model to \ac{FD}. \cite{Kuendee2019} uses a \ac{1D-CNN} to classify streaming sound data from a refrigerator compressor, achieving an accuracy of $99.4\%$, comparable to that of an \ac{ANN} model ($98.4\%$). \cite{Yan2022} applies an adaptive \ac{1D-CNN} model to raw sensor data from chillers augmented with synthetic faults. In the tests, \ac{1D-CNN} outperforms traditional algorithms such as \ac{SVM} and \ac{DT}. \ac{CNN} is used by \cite{Abhiraman2023, Soltani2020, Eom2019}. In \cite{Abhiraman2023}, the data are produced by simulating vaccine refrigerators through thermodynamic modelling in two settings. The first experiment considers six physical variables and the second one adds the ambient temperature. Considering supermarket refrigeration systems, the work in \cite{Soltani2020} simulates evaporator fan faults using two settings on the validation data to analyze the sensitivity of the \ac{CNN} model to the data quality: (1) with Gaussian noise and (2) with an offset-based perturbation. The evaluation shows that \ac{CNN} achieves an accuracy greater than $90\%$. The work in \cite{Eom2019} considers the simulation of a heat pump system with either heating or cooling mode and compares three architectures: a \ac{SNN}, a \ac{FCDNN} and a \ac{CNN} with ReLU as activation function and He normal \cite{he2015delving} as weight initialization. \ac{CNN} outperforms the other two architectures in detecting when the refrigerant charge amount is under, over or within a set limit.

\ac{RNN}-based approaches are proposed in several works. \ac{LSTM}, a type of \ac{RNN}, is used in \cite{Tian2021, Chen2023}. Considering the application of \ac{FD} to refrigerators, the work in \cite{Tian2021} introduces an \ac{LSTM} model configured with \ac{AutoML}. Through data normalization, \ac{mRMR} feature selection and \ac{GA}-based hyperparameter tuning, the \ac{LSTM} outperforms other models such as \ac{RNN} and \ac{SVM}, achieving a $91.63\%$ precision in detecting five faults with different \acp{SL}. The precision improves as fault severity increases. 
Thanks to the ability to learn complex temporal dependencies, \acp{DRNN} are applied to the \ac{FD} task. \cite{Taheri2021} compares different configurations of \acp{DRNN} with different layers and dropouts. The evaluation exploits two datasets (made available at \url{https://flexlab.lbl.gov/}) comprising data of single zone and multizone \acp{AHU} present in \ac{HVAC} systems, which are simulated with FlexLab. The tests compare hybrid models such as \ac{S-DRNN}, a \ac{DRNN} stacking multiple \ac{LSTM} units, and \ac{DTO-DRNN}, which adds a single \ac{MLP} hidden-to-output layer to summarize input information. The best result is given by the combination of \ac{S-DRNN} and \ac{DTO-DRNN}, which outperforms \ac{RF} and \ac{GB}.

The work in \cite{Chen2023} proposes a multi-stage approach and a hybrid \ac{LSTM}-based Siamese network. First, the training stage consists of training the Siamese network with a pair of time series subsequences to output 0 if they belong to different classes; otherwise, 1. Then, a subset of the train set (the support set) is randomly selected. Finally, the class of test time series subsequences is computed. Each test subsequence is paired with all the support set subsequences and given in input to the trained Siamese network. The average similarity with a test subsequence is computed for each class in the support set. The test subsequence class is the one with the highest average similarity. The proposed network outperforms the baseline method, a standard \ac{LSTM} model.

\paragraph{Evaluation}

\begin{figure}[]
  \centering
  \includegraphics[width=\textwidth]{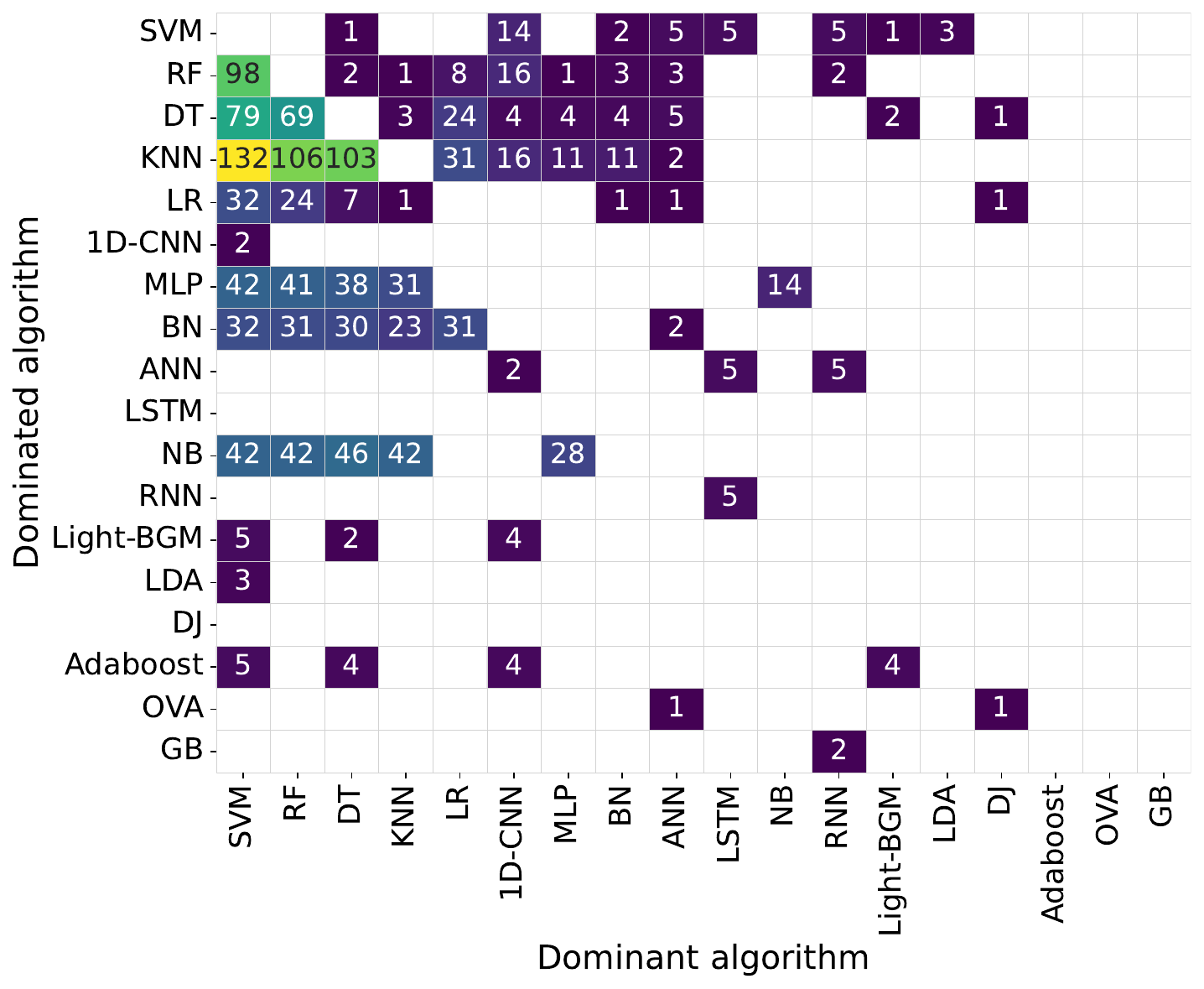}
  \caption{Visualization of pairwise comparisons of \ac{FD} algorithms. The x-axis reports the dominant algorithms, while the y-axis reports the dominated algorithms. Each cell reports the number of times that a dominant algorithm is found to outperform a dominated algorithm in the surveyed literature}
  \label{fig:fd heatmap}
\end{figure}

Given the high number of algorithms used in the surveyed \ac{FD} literature and the lack of a systematic comparison on the same dataset, we summarize the relative performances of the different techniques by visualizing the outcomes of the direct comparisons found in the reviewed works. Such representation is not a replacement for a head-to-head performance comparison on the same real-world dataset. However, it provides at least an impression of which algorithms have been employed more frequently for \ac{FD} and the number of tests in which a given algorithm has been found to outperform another.

Figure \ref{fig:fd heatmap} shows the algorithms used in the surveyed \ac{FD} works, reporting, in each cell, the number of times that a dominant algorithm has been found to outperform a dominated algorithm. The algorithms are sorted based on the number of times they are dominating other approaches in decreasing order of dominance. \ac{ML} algorithms are used and compared more often than \ac{DL} algorithms (80.87\% comparisons in total). Indeed, the 5 most used approaches are \ac{ML} algorithms (\ac{SVM}, \ac{RF}, \ac{DT}, \ac{KNN}, \ac{LR}). The single most used algorithm, \ac{SVM}, is compared mostly with other \ac{ML} algorithms and dominates 92.91\% of the time. It is often compared with \ac{KNN} and \ac{RF}, which are outperformed in most cases. This result stems from the ability of \ac{SVM} to perform non-linear classification by mapping inputs into high-dimensional feature spaces. This ability allows \ac{SVM} to be more robust to overfitting than \ac{KNN} when K is small \cite{bzdok2018machine}. Moreover, \ac{RF} struggles with many time-dependent features due to its bootstrap phase \cite{goehry2021random}.
\ac{SVM} also outperforms other methods such as \ac{LR} and \ac{DT} \cite{Yan2020}. \ac{LR} is a linear classifier and does not consider non-linear relationships between features, as \ac{SVM} does. Time-dependent data often exhibit non-linear behaviours, as shown in \cite{Fan2020, Yan2017, Dey2020}, making \ac{SVM} usually a better choice for \ac{FD}. \ac{DT}, similar to \ac{RF}, performs poorly in the presence of many features. In contrast, \ac{SVM} can handle the previous issues thanks to its regularizer and ability to deal with high-dimensional spaces, as shown in \cite{Yan2022, Yan2021, Zhu2021}. \ac{NB} never outperforms \ac{SVM} due to its strong assumption of not considering any correlations between features, as shown in \cite{Mukherjee2012}.

\ac{DL} methods are less used (19.13\% comparisons in total), notwithstanding their good performances in \cite{Zhu2021, Taheri2021, Tian2021}. Considering \ac{DL} approaches, they almost always outperform \ac{SVM}. The effectiveness of \ac{DL} methods on this task depends on their multiple hidden layers, which can capture relationships and patterns in the data \cite{Tian2021, Chen2023}. Indeed, \ac{1D-CNN}, \ac{LSTM}, \ac{RNN} and \ac{ANN} outperform the methods they are compared with in 96.77\%, 100.00\%, 73.68\% and 61.29\% of the cases, respectively. Moreover, \ac{1D-CNN}, \ac{LSTM}, \ac{RNN} surpass other \ac{ML} methods 96.43\%, 100.0\%, 100.0\% of the times, respectively. Notwithstanding the good results achieved by \ac{LSTM}, it is used by only two works \cite{Tian2021, Chen2023}, leading to fewer comparisons with respect to the most used \ac{ML} methods. Considering \ac{DL} algorithms only, \ac{1D-CNN} surpasses \ac{ANN} thanks to its filters that capture local dependencies and patterns. \ac{LSTM} achieves better results with respect to \ac{RNN} and \ac{ANN}. Specifically, \ac{LSTM} is a type of \ac{RNN} in which the vanishing gradient problem is mitigated \cite{Noh2021}. The ability of \ac{LSTM} to treat sequential data gives an advantage with respect to \acp{ANN} when dealing with time series, as \acp{ANN} are not designed to consider time-dependent patterns. Moreover, deep \acp{ANN} can also suffer from the vanishing gradient problem, as in the case of \acp{RNN}.

Figure \ref{fig:fd heatmap} also highlights that some of the top-performing algorithms (e.g., \ac{LSTM} and \ac{SVM}) are rarely compared to each other or are used only in a few researches. Further studies would be beneficial to highlight their advantages and disadvantages in datasets with different characteristics.

\subsubsection{Fault Prediction}
\label{sec:app_fp}

While \ac{FD} targets the identification of faults as they occur, \ac{FP} aims to predict future failures. 
%
Table \ref{tab:fault prediction results} summarizes the characteristics of the surveyed \ac{FP} works. In the case of \cite{Jiang2023}, "L+F" denotes the combination of \ac{LSTM-AE} and \ac{FFNN}. The body of research on \ac{FP} for compressor-based machines is more limited than on \ac{FD}. It considers machines in diverse settings (residential, industrial and other types of buildings), but the surveyed papers do not provide public datasets and only a few works employ real data. In general, faults are annotated using a rule-based approach for real data and through simulation for synthetic data.

\begin{sidewaystable}
  \centering
  \caption{Summary of the surveyed \ac{FP} works and of their characteristics. All approaches are supervised}
\label{tab:fault prediction results}

\begin{tabular*}{\textheight}{@{\extracolsep\fill}ccccccccc}
\toprule
\multicolumn{1}{c}{\textbf{Paper}} & \multicolumn{1}{c}{\textbf{Year}} & \multicolumn{1}{c}{\textbf{Real data}} & \multicolumn{1}{c}{\textbf{Field}} & \multicolumn{1}{c}{\textbf{Machine type}} & \multicolumn{1}{c}{\textbf{Public dataset}} & \multicolumn{1}{c}{\textbf{Features}} & \multicolumn{1}{c}{\textbf{\ac{GT}}} & \multicolumn{1}{c}{\textbf{Algorithm}} \\ \midrule
\multicolumn{1}{c}{{\cite{Kulkarni2018}}} & \multicolumn{1}{c}{2018} & \multicolumn{1}{c}{Y} & \multicolumn{1}{c}{Industrial} & \multicolumn{1}{c}{Refrigerator} & \multicolumn{1}{c}{N} & \multicolumn{1}{c}{40} & \multicolumn{1}{c}{R} & \multicolumn{1}{c}{\ac{RF}}\\ \midrule
{\cite{Bansal2021}} & \multicolumn{1}{c}{2021} & \multicolumn{1}{c}{Y} & \multicolumn{1}{c}{Domestic} & \multicolumn{1}{c}{Refrigerator} & \multicolumn{1}{c}{N} & \multicolumn{1}{c}{3} & \multicolumn{1}{c}{R} & \multicolumn{1}{c}{XGBoost} \\ \midrule
\multicolumn{1}{c}{{\cite{Padmanabh2021}}} & \multicolumn{1}{c}{2021} & \multicolumn{1}{c}{Y} & \multicolumn{1}{c}{Multistorey-Building} & \multicolumn{1}{c}{Chiller} & \multicolumn{1}{c}{N} & \multicolumn{1}{c}{72} & \multicolumn{1}{c}{R} & \multicolumn{1}{c}{\ac{KNN}} \\ \midrule
 \multicolumn{1}{c}{{\cite{Padmanabh2022}}} & \multicolumn{1}{c}{2022} & \multicolumn{1}{c}{Y} & \multicolumn{1}{c}{Buildings} & \multicolumn{1}{c}{\ac{HVAC}} & \multicolumn{1}{c}{N} & \multicolumn{1}{c}{5} & \multicolumn{1}{c}{R} & \multicolumn{1}{c}{\ac{SVM}} \\ \midrule
\multicolumn{1}{c}{{\cite{Jiang2023}}} & \multicolumn{1}{c}{2023} & \multicolumn{1}{c}{N} & \multicolumn{1}{c}{Commercial Buildings} & \multicolumn{1}{c}{Chiller} & \multicolumn{1}{c}{N} & \multicolumn{1}{c}{37} & \multicolumn{1}{c}{S} & \multicolumn{1}{c}{L+F} \\
 & & \multicolumn{1}{c}{N} & \multicolumn{1}{c}{Commercial Buildings} & \multicolumn{1}{c}{Chiller} & \multicolumn{1}{c}{N} & \multicolumn{1}{c}{42} & \multicolumn{1}{c}{S} & \multicolumn{1}{c}{L+F} \\ 
 & & \multicolumn{1}{c}{N} & \multicolumn{1}{c}{Commercial Buildings} & \multicolumn{1}{c}{Chiller} & \multicolumn{1}{c}{N} & \multicolumn{1}{c}{42} & \multicolumn{1}{c}{S} & \multicolumn{1}{c}{L+F} \\ 
& & \multicolumn{1}{c}{N} & \multicolumn{1}{c}{Commercial Buildings} & \multicolumn{1}{c}{Chiller} & \multicolumn{1}{c}{N} & \multicolumn{1}{c}{37} & \multicolumn{1}{c}{S} & \multicolumn{1}{c}{L+F} \\
 & & \multicolumn{1}{c}{N} & \multicolumn{1}{c}{Commercial Buildings} & \multicolumn{1}{c}{Chiller} & \multicolumn{1}{c}{N} & \multicolumn{1}{c}{37} & \multicolumn{1}{c}{S} & \multicolumn{1}{c}{L+F} \\ 

\bottomrule
\end{tabular*}

\end{sidewaystable}

The work in \cite{Bansal2021} uses XGBoost to predict faults in data collected by \ac{IoT} devices in refrigerators. First, data are preprocessed by combining features and removing invalid samples. Then, the model is trained using historical data aggregated daily. Despite achieving a precision of 91.4\%, the proposed method is limited to a three-day \ac{RW} and a single fault type.
\cite{Kulkarni2018} uses a multi-step approach for learning temperature signal patterns correlated to a faulty state and enabling the prediction of impending failures. The overall methodology is divided into three phases. The first step is score computation, which calculates derived quantities from input temperature signals. The second step is feature calculation, which aggregates the scores to obtain new features (e.g., by averaging some of the scores). The third step is model learning and consists of training an \ac{RF} binary classifier on the new features to predict failures. The classifier is trained on $\approx$ 70\% of the refrigerators and tested on the remaining $\approx 30\%$. The high precision and recall indicate its effectiveness, even though it uses only one sensor.

The work in \cite{Padmanabh2021} aims at predicting faults in chillers in multi-storey buildings based on sensor alarms. An alarm is generated whenever a sensor value exceeds its predefined limit and a combination of alarms might precede a failure.
Three \ac{ML} methods (\ac{KNN}, \ac{SVM} and \ac{LR}) are used to predict alarms and alarm predictions are combined to predict faults. The results of the evaluation show that \ac{KNN} outperforms \ac{SVM} and \ac{LR} in terms of accuracy, precision, recall and computational time.

The work in \cite{Jiang2023} predicts faults in chillers of commercial buildings. It uses the real time series of five different appliances, each comprising 40 features. An autoencoder builds a latent representation of the time series and a classifier takes such latent representation in input to predict faults that may be caused by low evaporator pressure, such as motor current overload and high oil temperature. The study compares three feature extraction architectures (\ac{FF-AE}, \ac{LSTM-AE} and \ac{PCA}) and three classifiers (\ac{FFNN}, Gaussian Model and XGBoost). In terms of accuracy, the most robust and high-performance models employ the \ac{FFNN} as the classifier. The combination of \ac{LSTM-AE} and \ac{FFNN} performs similarly to that of \ac{PCA} and \ac{FFNN}, but the latter has better accuracy in most experiments.


The work in \cite{Padmanabh2022} considers more than 60 different \ac{HVAC} systems, for which hundreds of features are measured, to predict faults at least 10 hours in advance. The datasets span over three years and are split by considering 2 years as the training set and 1 year as the testing set. First, five features are extracted and then \ac{ANN}, \ac{LR}, \ac{KNN} and \ac{SVM} are compared as fault predictors. Performances are evaluated using accuracy, precision, recall and $F_1$ score. \ac{SVM} achieves the best results for all the metrics. 

\paragraph{Evaluation}

\begin{figure}[]
  \centering
  \includegraphics[width=\textwidth]{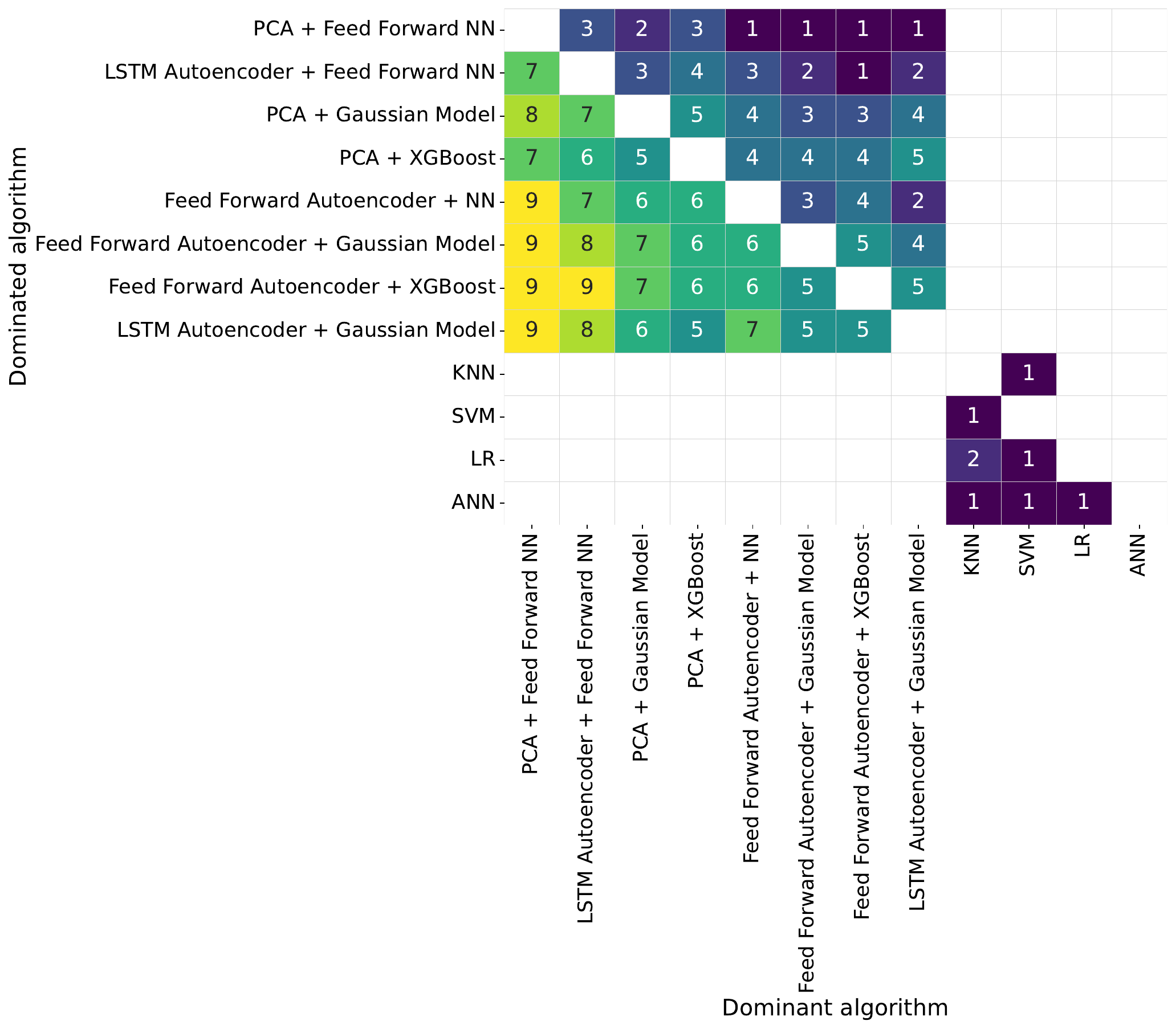}
  \caption{Visualization of pairwise comparisons of \ac{FP} algorithms. The x-axis reports the dominant algorithms, while the y-axis reports the dominated algorithms. Each cell reports the number of times that a dominant algorithm is found to outperform a dominated algorithm in the surveyed literature}
  \label{fig:fp heatmap}
\end{figure}

Figure \ref{fig:fp heatmap} shows the algorithms used in the surveyed \ac{FP} works, reporting, in each cell, the number of times that a dominant algorithm is found to outperform another approach. The algorithms are sorted in decreasing order of dominance. Compared with the \ac{FD} case, fewer comparisons between methods are found in the literature, most of which come from a single work \cite{Jiang2023}. The surveyed \ac{FP} approaches can be divided into two main groups: \ac{ML} (e.g. \ac{SVM}, \ac{LR} and \ac{KNN}) and \ac{DL} (e.g. \ac{LSTM} and \ac{FFNN}). 

In the realm of \ac{ML} methods, \ac{KNN} and \ac{SVM} achieve similar performances and the best option depends on the application and the dataset \cite{kone2018performance}. \ac{KNN} captures complex patterns and its output is more challenging to interpret \cite{bzdok2018machine}. Both methods outperform \ac{LR}, as shown, e.g., in \cite{Padmanabh2022}. 

In the realm of \ac{DL} methods, different architectures are compared in \cite{Jiang2023}. Specifically, the feature extractors are built by using \ac{PCA}, \ac{LSTM-AE} and \ac{FF-AE}. The classifiers comprise XGBoost, Gaussian Model and \ac{FFNN}. The \ac{LSTM-AE} combined to \ac{FFNN} on average outperforms \ac{PCA} + \ac{FFNN}, \ac{PCA}+Gaussian Model, \ac{PCA} + XGBoost, \ac{FF-AE} + \ac{FFNN}, \ac{FF-AE} + Gaussian Model, \ac{FF-AE} + XGBoost and \ac{LSTM-AE} + Gaussian Model. 
This result depends on the greater robustness of the \ac{LSTM-AE} + \ac{FFNN} combination with respect to the data distribution drifts from the training to testing data \cite{Jiang2023}. Moreover, \ac{LSTM-AE} exhibits better ability to capture long-term patterns and to mitigate the vanishing gradient issue. 

Furthermore, the evaluation of \ac{FP} in this context often lacks comparisons with baseline approaches. For example, the work in \cite{Kulkarni2018} does not include comparisons of \ac{RF} with any other methods. Moreover, there are no direct comparisons between \ac{ML} and \ac{DL} methods.

\subsubsection{Forecasting}
\label{sec:app_forecasting}

Forecasting is the process of predicting the values of variables of interest based on past data. It has several applications for compressor-based machines, e.g., to prevent energy waste in refrigerators and \ac{HVAC} systems \cite{SalaCardoso2018, Qian2020, Yu2021}.

\ac{ML} algorithms are widely used in forecasting. Examples are \ac{GPR} \cite{Moradkhani2022, Ahmad2019} and \ac{SVR} \cite{LeCam2017}. \ac{DL} approaches are also employed, e.g., \ac{ANN} \cite{Kim2020} and \ac{RNN}-based architectures \cite{ChaerunNisa2021, Xie2023, Brusokas2021, Li2022, Yu2021, Xie2020, Petroanu2020, Xu2021, Xu2019_Indoor, Liu2023}. 



Table \ref{tab:forecasting_datas_and_results} summarizes the characteristics of the surveyed works. For \cite{Xu2019_GSHPS, Xu2019_Indoor}, the first row refers to the one-step forecasting, while the second refers to the multi-step forecasting. The best algorithm of \cite{LeCam2017} is indicated as \ac{SVR}-C, where `C' indicates the type of preprocessing (see the paper for the details). 

\begin{sidewaystable}

\caption{An overview of forecasting works and their characteristics. All approaches are supervised}
\label{tab:forecasting_datas_and_results}
\centering
\begin{tabular*}{\textheight}{@{\extracolsep\fill}ccccccccc}

\toprule
{\textbf{Paper}} & \textbf{Year} & \textbf{Real data} & \textbf{Field} & \textbf{Machine type} & \textbf{Public dataset} & \textbf{Features} & \textbf{Algorithm} \\ \midrule
{\cite{LeCam2017}} & 2017 & Y & University Building & \ac{HVAC} & N & 15 & {\color[HTML]{333333} \acs{SVR}-C} \\ \midrule

{\cite{SalaCardoso2018}} & 2018 & Y & Office & Heat pump and Chiller & N & 7 & {\color[HTML]{333333} \acs{RNN}-\acs{ANFIS}} \\ \midrule

{\cite{Ahmad2019}} & 2019 & N & Office & Heat pump & N & 11 & {\color[HTML]{333333} \acs{GPR}} \\
& & & & & & & \acs{LMB-NN} \\ \midrule
{\cite{Xu2019_Indoor} } & 2019 & Y & Buildings & Heat pump & N & 1 & {\color[HTML]{333333} Correction-Error \acs{LSTM}} \\
&  & & & & & & Correction-Error \acs{LSTM} \\ \midrule
{\cite{Xu2019_GSHPS}} & 2019 & Y & Office Building & Heat pump & N & 1 & {\color[HTML]{333333} \acs{VMD}-\acs{PSO}-\acs{ELM}} \\
& & & & & & &  {\color[HTML]{333333} \acs{VMD}-\acs{PSO}-\acs{ELM}} \\ \midrule
{\cite{Kim2020}} & 2020 & Y & Office Building & \acs{AHU} & N & 5 & {\color[HTML]{333333} \acs{NARX-NN}} \\
& &  Y & Office Building & Chiller & N & 5 & {\color[HTML]{333333} \acs{NARX-NN}} \\ \midrule
{\cite{Xie2020}} & 2020 & N & Industrial & Heat pump & N & 5 & {\color[HTML]{333333} \acs{LSTM}-\acs{BPNN}+\acs{BPNN}} \\ \midrule
{\cite{Petroanu2020} } & 2020 & Y & Commercial center & Refrigerator & Y\footnote{\url{https://www.mdpi.com/2071-1050/13/1/104/s1}} & 1 & {\color[HTML]{333333} \acs{Bi-LSTM}-\acs{FITNET}} \\ \midrule
{\cite{Faria2020}} & 2020 & Y & Office building & Refrigerator & N & 4 & {\color[HTML]{333333} \acs{WM}} \\ \midrule
{\cite{Qian2020}} & 2020 & Y & Industrial & \ac{HVAC} & N & 3 & {\color[HTML]{333333} TrAdaboost(\acs{SVR})} \\
& & & & & & & {\color[HTML]{333333} \ac{ANN}} \\ \midrule
{\cite{ChaerunNisa2021}} & 2021 & Y & University Building & Chiller & N & 6 & {\color[HTML]{333333} \acs{LSTM}} \\ \midrule
{\cite{Yu2021} } & 2021 & Y & Office Building & Chiller & N & 9 & {\color[HTML]{333333} \acs{Seq2Seq LSTM}} \\ \midrule

{\cite{Brusokas2021}} & 2021 & N & Residential buildings & Heat pump & Y\footnote{\url{https://pages.nist.gov/netzero/data.html}} & 3 & {\color[HTML]{333333} \acs{LSTM}} \\
{\color[HTML]{000000} } & & N & Residential buildings & Heat pump & Y\footnote{\label{NYSERDA}\url{https://www.nyserda.ny.gov/}} & 3 & {\color[HTML]{333333} \acs{LSTM}} \\
{\color[HTML]{000000} } & & N & Residential buildings & Heat pump & Y\footref{NYSERDA} & 3 & {\color[HTML]{333333} \acs{LSTM}} \\ \midrule
{\cite{Xu2021}} & 2021 & Y & University Building & Heat pump and Chiller & N & 6 & {\color[HTML]{333333} \acs{A-LSTM}} \\ \midrule
{\cite{Moradkhani2022} } & 2022 & N & Industrial & Other & Y\footnote{Available on request} & 15 & {\color[HTML]{333333} \acs{GPR}} \\ \midrule
{\cite{Li2022}} & 2022 & Y & Industrial & Chiller & N & 19 & {\color[HTML]{333333} \ac{LSTM}} \\
&  & Y & Industrial & Chiller & N & 19 & {\color[HTML]{333333} \acs{LSTM}} \\
&  & Y & Industrial & Air Conditioner & N & 19 & {\color[HTML]{333333} \acs{LSTM}} \\ \midrule

{\cite{Liu2023}} & 2023 & Y & Office Building & \ac{HVAC} & N & 11 & {\color[HTML]{333333} \acs{RF}-\acs{ISSA}-\acs{LSTM}} \\
&  & Y & School Building & \ac{HVAC} & N & 11 & {\color[HTML]{333333} \ac{RF}-\acs{ISSA}-\acs{LSTM}} \\
&  & Y & School Building & \ac{HVAC} & N & 11 & {\color[HTML]{333333} \ac{RF}-\acs{ISSA}-\acs{LSTM}} \\ \midrule

{\cite{Xie2023}} & 2023 & N & Industrial & Heat pump & N & 7 & \ac{LSTM} \\ 
\bottomrule

\end{tabular*}

\end{sidewaystable}

In contrast to \ac{FD} and \ac{FP}, a higher percentage of experiments ($\approx 67\%$) use real data because the forecasting task does not require labour-intensive data annotations. However, only 26\% of the surveyed works employ publicly available datasets and none of the used datasets is both public and built from real data. The employed time series typically comprise fewer features ($\approx 8$) compared to \ac{FD} ($\approx 36$) and \ac{FP} ($\approx 39$) and more features compared to \ac{CPD} ($\approx 4$). Forecasting is applied to 6 machine types, compared to the 4 types mentioned in the \ac{FD} works, notwithstanding the comparable amount of surveyed works, which suggests that forecasting is a task of interest for a broader range of systems.

The work in \cite{Faria2020} combines \ac{WM} to forecast the load consumption in the refrigerator using 4 input variables. \ac{WM} is trained to generate fuzzy rules (i.e. conditional statements imitating the reasoning of experts), as in \cite{casillas2000improving}. \ac{WM} is compared to \ac{SVM} to forecast the load consumption of refrigerators with a \ac{PW} of 24 hours. \ac{WM} achieves a better \ac{MAE} ($22.54$ W) compared to \ac{SVM} ($42.62$ W).

The work in \cite{LeCam2017} combines \ac{SVR} with several preprocessing methods to forecast five target variables for the \ac{HVAC} system of a 5-level University building: supply air flow rate and electric demand fans for the \acp{AHU}, cooling coil load and electric demand of chillers and whole building cooling load. Preprocessing selects regressors (i.e., input features for \ac{SVR}) and clusters (i.e., typical daily profiles of the target variables). Three preprocessing methods are compared. Method `A' implements a forecasting model for each cluster and target variable using the regressors from the whole dataset. Method `B' is analogous to `A' but uses only regressors specific to that cluster. Method `C' creates a unique model for each target variable using regressors from the whole dataset. The preprocessing output is fed as input to an \ac{SVR}, which forecasts all the target variables with a horizon of 15 minutes on the test set. The \ac{SVR} with preprocessing `C' (\ac{SVR}-C) has the lowest \ac{RMSE} on the total electric demand forecasting of the building. Other preprocessing methods are less effective because they do not exploit information from different daily profiles. 

The work in \cite{SalaCardoso2018} introduces a hybrid model composed of an \ac{RNN} and an \ac{ANFIS} to forecast the thermal power demand in heat pumps, \acp{AHU} and chillers in buildings. The \ac{RNN} is trained using 11 variables to learn activity indicators (the occupancy statuses in the building), which are correlated with energy usage. The activity indicators and other variables (e.g., temperature) are fed as input to the \ac{ANFIS}, a neural network that learns fuzzy rules. An ablation study shows that using only activity indicators or only the other variables leads to worse results for all the analyzed metrics (\ac{RMSE}, \ac{MAPE}, \ac{MAE}, \ac{MAX}, $R^2$ and \ac{TIME}).

The work in \cite{Moradkhani2022} forecasts the \ac{HTC} in helically coiled tubes using a public simulated dataset with 15 variables. The comparison between \ac{LSFM} \cite{xin2009xiao}, \ac{GP}, 15 existing methods and \ac{ML} methods (\ac{MLP}, \ac{GPR} and \ac{RBF} networks) shows that \ac{ML} methods achieve better performances. \ac{GPR} achieves the best \ac{AARE} ($5.93\%$) thanks to its ability to estimate the complex and non-linear behaviour of \ac{HTC}. Existing methods' performances are significantly worse than those of novel approaches (\ac{AARE} from 23.05\% to over 1000\%).

The work in \cite{Ahmad2019} uses \ac{GPR} for forecasting the short-term (1 week) and long-term (1 month) cooling load demand of the water source heat pump in a simulated office building. The dataset comprises 11 variables (e.g., weather parameters and energy consumption of the water source heat pumps). Three \ac{ML} algorithms are compared: \ac{MLR}, \ac{GPR} and \ac{LMB-NN}. The latter is a \ac{BPNN} improved by the Levenberg–Marquardt algorithm, which can solve non-linear least squares problems efficiently. For long-term forecasting experiments, \ac{LMB-NN} achieves better performance (in terms of $R$, \ac{MAE}, \ac{MAPE} and the relative standard deviation CV) due to its ability to find extended patterns, while \ac{GPR} outperforms the other approaches for short-term forecasting experiments.

The work in \cite{Kim2020} uses \ac{NARX-NN}, an \ac{FFNN} with a hidden layer, to predict energy consumption in a chiller and \ac{AHU} installed in an office building air conditioner systems. \ac{NARX-NN} takes in input operation conditions of equipment, seasonality data and historical energy input data. Performances are evaluated using \ac{CV(RMSE)} and \ac{MBE}, metrics that satisfy the \ac{ASHRAE} guidelines about performance evaluation \cite{guideline2014measurement}. Best results are obtained for long-term forecasting.

The work in \cite{Qian2020} uses an \ac{ANN} and TrAdaboost(\ac{SVR}) to forecast heating and cooling load in an energy centre \ac{HVAC} system. The dataset contains four variables: humidity, temperature, heating load demand and cooling load demand. \ac{ANN} is used to forecast the short-term (daily to weekly) and medium-term (monthly to annual) cooling and heating load. TrAdabbost(\ac{SVR}) forecasts the target cooling and heating load for the next year using simulated training data and a small amount of actual data. TrAdaboost (\ac{SVR}) achieves higher accuracy than \ac{ANN} as the training set size reduces.


The work in \cite{ChaerunNisa2021} compares \ac{MLP}, \ac{LSTM} and \ac{1D-CNN} to forecast the next-minute power consumption in a water-cooled University building chiller. The dataset consists of 6 variables collected over 13 days. The three algorithms are trained using different training set lengths (10, 11 and 12 days). \ac{LSTM} performances are better for the 12 days training set (\acs{RMSE} of $\approx 1.4 kW$) than the other two methods (the \ac{RMSE} of \ac{MLP} is $\approx 2.6 kW$ and the one of \ac{1D-CNN} is $\approx 1.5 kW$). Longer training sets are found to produce lower \ac{RMSE} values.

The work in \cite{Yu2021} uses \ac{Seq2Seq LSTM} (i.e., an encoder-decoder architecture where \ac{LSTM} is part of both the encoder and the decoder) to predict the next 3 days \ac{COP} in five water-cooled systems installed in an office building. The dataset comprises 9 variables (e.g., the evaporator and the condenser discharge temperatures). \ac{Seq2Seq LSTM} outperforms \ac{LSTM}, V-\ac{LSTM}, ResRNN, \ac{RNN} and \ac{GRU} in terms of \ac{MAE} ($0.109$, compared to $0.121$, reached by the second-best algorithm).

The work in \cite{Xie2023} uses \ac{LSTM} to forecast the heating load of a \ac{HGSHP}. A \ac{GSHPS} is a heating and cooling system that transfers heat from the ground to provide energy for buildings. These systems may deteriorate due to thermal imbalance. 
The dataset is simulated and comprises 6 features (power consumption, inlet/outlet water temperature on the load side, inlet/outlet water temperature on the source side and flow rate on the load/source side). \ac{LSTM} performs a multi-step forecast with varying \acp{PW} (1 to 10 hours) and \acp{RW} (5 to 30 hours). The optimal \ac{RW}-\ac{PW} combination is 25 hours and 5 hours, respectively. The \ac{MAE} and \ac{MAPE} decrease by 20.6\% and 23.6\% respectively compared to the case of a 10 hours \ac{PW} and a 5 hours \ac{RW}.

The work in \cite{Brusokas2021} uses \ac{LSTM} to forecast the indoor temperature in 3 simulated residential buildings datasets that comprise indoor temperature, outdoor temperature and power consumption of heat pumps. Forecasting performances are evaluated using \ac{RMSE}, \ac{MAE} and \ac{MAPE}. \ac{LSTM} is compared with 3 other methods (\ac{GRU}, \ac{LR} and \ac{SARIMAX}). \ac{LSTM} and \ac{GRU} outperform \ac{LR} and \ac{SARIMAX} because they can better capture long-term dependencies in sequential data. Tuning of \ac{LSTM} hyperparameters leads to slightly better performances than \ac{GRU}.

The work in \cite{Xie2020} uses \ac{LSTM}-based hybrid models to forecast the cooling load, sensible heat and latent heat of heat pumps installed in an air conditioner. The dataset is simulated and contains five variables. This work proposes two novel hybrid methods to make a 1-hour prediction: \ac{LSTM}-\ac{BPNN} (\ac{LSTM} combined with a \ac{BPNN}) and \ac{LSTM}-\ac{BPNN} + \ac{BPNN} (\ac{LSTM} combined with two \acp{BPNN}). Performances are evaluated using \ac{RMSE} and \ac{MAPE}. The first method outperforms \ac{SVM}, \ac{BPNN}, \ac{LSTM} and \ac{LSTM}-\ac{BPNN}+\ac{BPNN} in forecasting the sensible heat and \ac{BPNN} has the best performance in predicting latent heat and the cooling load.

The work in \cite{Petroanu2020} uses an \ac{LSTM}-based architecture to forecast, among others, the energy consumption of a refrigerator room, with a \ac{PW} of one month and a \ac{RW} of one hour. The dataset contains real time series of refrigerator power consumption. The work employs a \ac{Bi-LSTM} whose input is constituted by delayed input sequences. Then, FITNET takes in input \ac{Bi-LSTM} output to make predictions. This architecture achieves good performance (with a \ac{MSE} ranging from $\approx 0.0003$ to $\approx 0.0007$) due to the forward and backward propagation of the \ac{Bi-LSTM} component and the capacity of \ac{FITNET} to forecast accurately in the context of a fast-training process.

The work in \cite{Xu2019_GSHPS} uses modal decomposition-based ensemble learning to forecast the next 5 to 30 minutes energy consumption of \acp{GSHPS} in an office building. The proposed approach comprises three components. \ac{VMD} decomposes the time series into oscillatory components, \ac{ELM} forecasts the energy consumption and \ac{PSO} is used for \ac{ELM} hyperparameter selection. A comparison of \ac{VMD}  with \ac{EMD} highlights the effectiveness of \ac{VMD} in decomposing non-linear data. The assessment of \ac{ELM} and \ac{SVM} shows that \ac{SVM} requires more time for processing a large number of samples. Performances are evaluated using \ac{RMSE} for several techniques 
 and combinations: \ac{BPNN}, \acs{BOA}-\ac{SVM}, \ac{VMD}-\ac{PSO}-\ac{ELM}, \ac{VMD}-\acs{BOA}-\ac{SVM}, \ac{EMD}-\ac{BPNN} and \ac{EMD}-\acs{BOA}-\ac{SVM}. The top-performing approach in both one-step and multi-step forecasting is \ac{VMD}-\ac{PSO}-\ac{ELM}, with an \ac{RMSE} of $\approx 9.1$ and $\approx 23.4$ respectively.

\paragraph{Evaluation}
\begin{figure}[h]
  \centering
  \includegraphics[width=\textwidth]{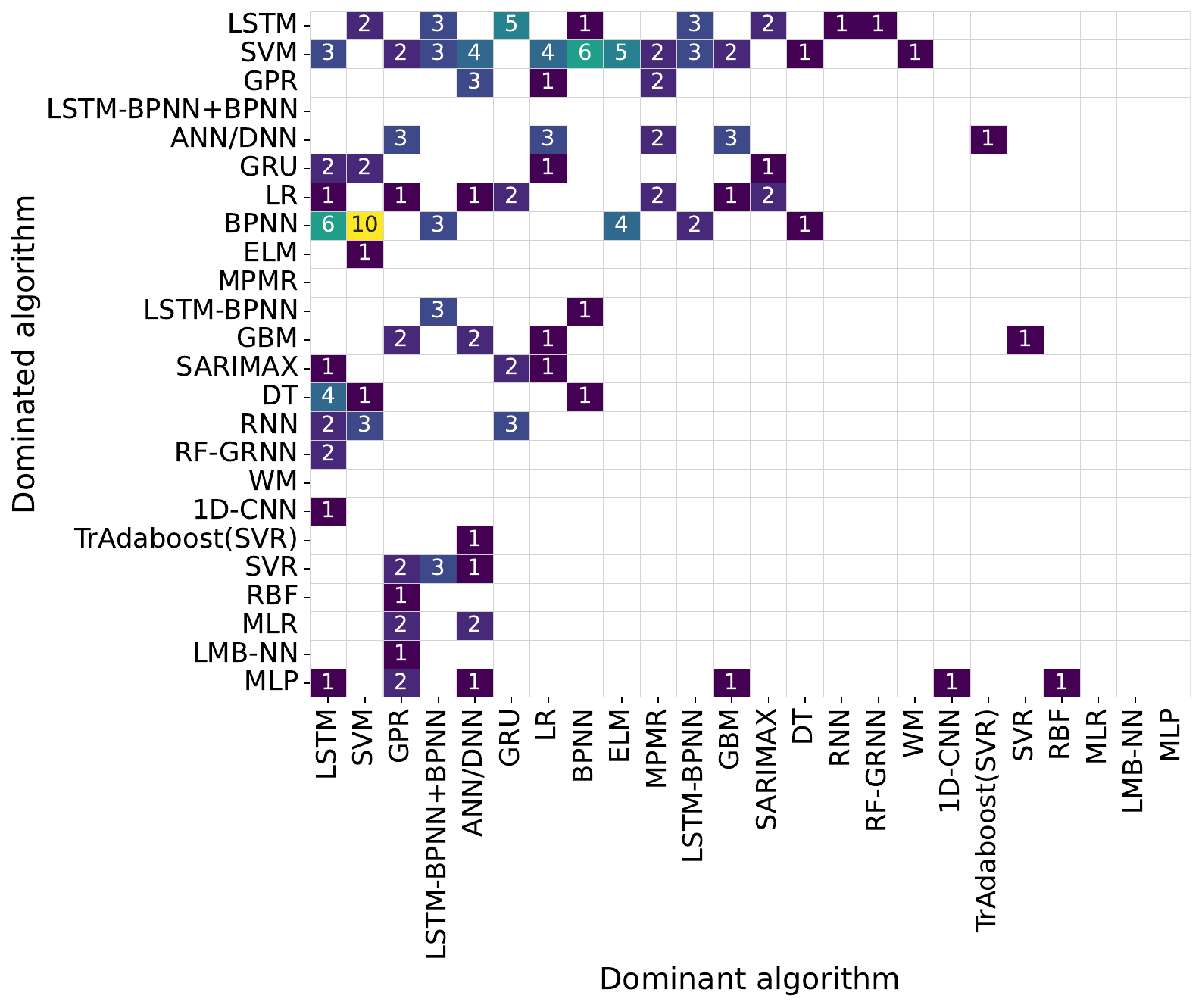}
  \caption{Visualization of pairwise comparisons of forecasting algorithms. The x-axis reports the dominant algorithms, while the y-axis reports the dominated algorithms. Each cell reports the number of times that a dominant algorithm is found to outperform a dominated algorithm in the surveyed literature}
  \label{fig:forecasting heatmap}
\end{figure}

Figure \ref{fig:forecasting heatmap} shows the algorithms used in the surveyed forecasting works, reporting, in each cell, the number of times that a dominant algorithm is found to outperform another approach. The algorithms are sorted in decreasing order of dominance. The surveyed methods can be divided into three main groups: \ac{ML} (e.g. \ac{SVM}, \ac{LR} and \ac{SVR}), statistical (e.g., \ac{SARIMAX}) and \ac{DL} methods (e.g. \ac{LSTM} and \ac{RNN}). For this task, \ac{ML} methods are used 39.4\% of the times, \ac{DL} methods 57.6\% and other methods (e.g., \ac{SARIMAX} and \ac{WM}) 3\% of the times.

Vanilla \acp{RNN} are not widely used because of the vanishing gradient problem, mitigated by other \ac{RNN}-based architectures, such as \ac{LSTM} and \ac{GRU}. \ac{LSTM} outperforms several \ac{ML} algorithms in 80\% of the cases. In particular, \ac{SVM} is surpassed $\approx$ 60\% of the times and \ac{DT}, \ac{GPR}, \ac{GBM}, \ac{SVR}, TrAdaboost (\ac{SVR}), \ac{MPMR} and \ac{LR} 100\% of the times. Similar to the \ac{FD} case, this result suggests that \ac{LSTM} is an effective method, as it can effectively capture long-term temporal dependencies and handle sequential data effectively, making it a suitable choice for forecasting tasks. In addition, \ac{LSTM} surpasses \ac{BPNN} in 85.7\% of the cases thanks to its ability to model time dependencies. The work in \cite{Xie2020} suggests combining the advantages of \ac{LSTM} with the ones of \ac{BPNN} to forecast load profiles, as this combination can exploit the ability of \ac{LSTM} to capture long-term dependencies in data and the ability of \ac{BPNN} to deal with data without temporal dependencies.

\ac{GRU} achieves superior performances with respect to \ac{LSTM} 71.4\% of the times in the surveyed works. The work in \cite{Rau2021} shows that \ac{LSTM} performance increases for large amounts of training data. In general, \ac{GRU} and \ac{LSTM} achieve contrasting results \cite{Rau2021, 9073069, Touzani2021, 9936434}. Instead, when compared to \ac{LR} and \ac{SARIMAX}, \ac{GRU} and \ac{LSTM} can handle and learn from complex, non-linear and long data sequences. 

\ac{SVM} achieves similar performance to the one gained by \ac{DT} \cite{Xu2019_Indoor} and outperforms \ac{BPNN} 62.5\% of the times \cite{Xu2019_Indoor, Xu2019_GSHPS} because it can project data into higher-dimensional spaces and deal with non-linear data. However, \ac{SVM} is consistently outperformed by \ac{GPR}, \ac{MPMR} and \ac{GBM} because of their suitability for modelling complex and uncertain data such as heating and cooling load data.

\subsubsection{Change Point Detection}
\label{sec:app_cpd}
\ac{CPD} aims to identify when a time series's behaviour or underlying distribution changes. This task is useful in detecting shifts in time series to identify changes in the working regime of machines that may cause faults or abnormal behaviours \cite{van2018detection, Pereira2018}. Statistical, \ac{ML} and \ac{DL} methods are used to address this task.

Table \ref{tab:change point detection data and results} reports the characteristics of the surveyed \ac{CPD} research. There are fewer works than those addressing the other tasks and none of them uses public datasets. Moreover, the number of features is usually smaller than the other tasks, suggesting that a limited number of variables is sufficient for detecting change points.

\begin{sidewaystable}

\centering
\caption{An overview of \ac{CPD} works and their characteristics}
\label{tab:change point detection data and results}

\begin{tabular}{p{0.09\textwidth}cccccccccc}

\toprule
\textbf{Paper} & \textbf{Year} & \textbf{Real data} & \textbf{Field} & \textbf{Machine type} & \textbf{Public dataset} & \textbf{Features} & \textbf{\ac{GT}} & \textbf{Supervision} & \textbf{Algorithm} \\ \midrule
\cite{van2018detection} & 2018 & Y & Commercial Shop & Refrigerator & N & 3 & N & U & \acs{ANN} \\ \midrule
\cite{Pereira2018} & 2018 & Y & Domestic & Heat Pump & N & 5 & E & U & \acs{CPA} \\
& & Y & Domestic & Heat Pump & N & 5 & E & U & \acs{CPA} \\ \midrule
\cite{Touzani2019} & 2019 & N & Primary School & Chiller & N & 4 & S & S & \acs{GBM}-Euclidean-\acs{PELT} \\ 
 & & N & Primary School & Chiller & N & 4& S & S & \acs{GBM}-\acs{CORT}-\acs{PELT} \\ 
 & & N & Local Office & Chiller & N & 3 & S & S & \acs{GBM}-\acs{CORT}-\acs{PELT} \\
& & N & Local Office & Chiller & N &3 & S & S & \acs{GBM}-Euclidean-\acs{PELT} \\
& & N & Local Office & Chiller & N & 3 & S & S & \acs{GBM}+\acs{CORT}-\acs{PELT} \\

\bottomrule
\end{tabular}

\end{sidewaystable}

The work in \cite{Touzani2019} uses \ac{ML} and statistical methods to analyse the energy consumption 
time series of school and office buildings and detect events such as the change of the electrical base load of the building. It uses data from 8 simulated buildings, 4 with chillers and 4 without compressor-based machines. The latter are not considered in this survey. The datasets comprise 2 to 3 additional variables: time of week, temperature and, when the considered building is a school, a vacation indicator. The proposed method (\acs{GBM}-\acs{CORT}-\acs{PELT}) combines \ac{GBM}, \ac{CORT} and \ac{PELT}. \ac{GBM} is an ensemble tree-based \ac{ML} algorithm that generates a model of the energy consumption using the input variables. It predicts the energy consumption for a \ac{PW} called post period. \ac{CORT} \cite{chouakria2007adaptive} computes the dissimilarity between the \ac{GT} and the \ac{GBM} prediction. \ac{PELT} \cite{Killick2012} is a \ac{CPD} algorithm that detects changepoints on \ac{GBM} output. \acs{GBM}-\acs{CORT}-\acs{PELT} is compared with \acs{GBM}-Euclidean-\acs{PELT}, which replaces \ac{CORT} dissimilarity with Euclidean dissimilarity, and \ac{PELT}, which is applied directly on input data. \ac{GBM}-\ac{CORT}-\ac{PELT} achieves the highest precision for 3 datasets (33.3\% to 66.7\%, with a mean of $\approx 56\%$ and a median of $\approx 62\%$). For instance, one of the primary schools achieves a precision of 66.7\%, which is superior to the one obtained using the second algorithm (37.5\%) and the third algorithm (0.0\%). \ac{GBM}-Euclidean-\ac{PELT} achieves the highest precision for 2 datasets (of which one tied with \ac{GBM}-\ac{CORT}-\ac{PELT}) and \ac{PELT} is always surpassed.


The work in \cite{van2018detection} aims to detect change points that occur when a refrigerator transitions from an active to an inactive state and vice versa and is an intermediate step for the task of energy consumption disaggregation. It uses a real-world, unlabelled dataset comprising a commercial shop's aggregated power consumption time series as input. For validation, it exploits the individual power consumption time series of three refrigeration units in the shop: a display cabinet, an under-counter fridge and an upright fridge.
The authors compare two approaches (clustering-based and \ac{ANN}-based).
The clustering-based approach performs \ac{CPD} using an event detection algorithm to find refrigeration cycles in the aggregated signal. First, the signal is denoised; then, it is partitioned into cycles delimited by rising and falling edges. Rising edges are clustered based on their magnitudes, using Euclidean distance and a number of clusters equal to the number of refrigeration units. In the evaluation phase, every detected rising edge is validated by comparing the aggregate time series and the signals acquired from the individual units. A detection is validated when the cycle's rising and falling edge occur within a 10-second window with respect to the unit's time series. 
The \ac{ANN} approach splits the aggregated power consumption into overlapping windows with a 1-sample shift. Each window is used to estimate the power consumption of a refrigeration unit at its centre point. The network is trained by minimizing the \ac{MSE} between the actual and the predicted power consumption. In the evaluation phase, the k-means algorithm is applied to the time series of each refrigeration unit to identify two clusters, one for active states and the other for inactive states. The mean of the two cluster centres is used as a threshold to detect cycles. \ac{ANN} shows better precision and recall in detecting change points (up to $+13\%$ precision and $+9\%$ recall) w.r.t. the clustering method.

The work in \cite{Pereira2018} uses \ac{CPD} to characterize the actions of residential buildings' occupants. The possible actions include using the heating system, cooking, showering and opening the windows. Sensors collect values for 5 input variables: temperature (external and internal), relative humidity (external and internal) and $CO_2$. The change point of the heating system (switch on and off) is detected as a user's action. The \ac{GT} is collected using daily journals filled manually and sensors. \ac{CPD} is performed using the \ac{CPA} algorithm, as implemented in the R package \texttt{changepoint}. The \ac{GT} is compared with \ac{CPA} predictions and performances are evaluated using accuracy-based metrics, showing the effectiveness of the proposed approach.

\paragraph{Evaluation}

\begin{figure}[h]
  \centering
  \includegraphics[width=.8\textwidth]{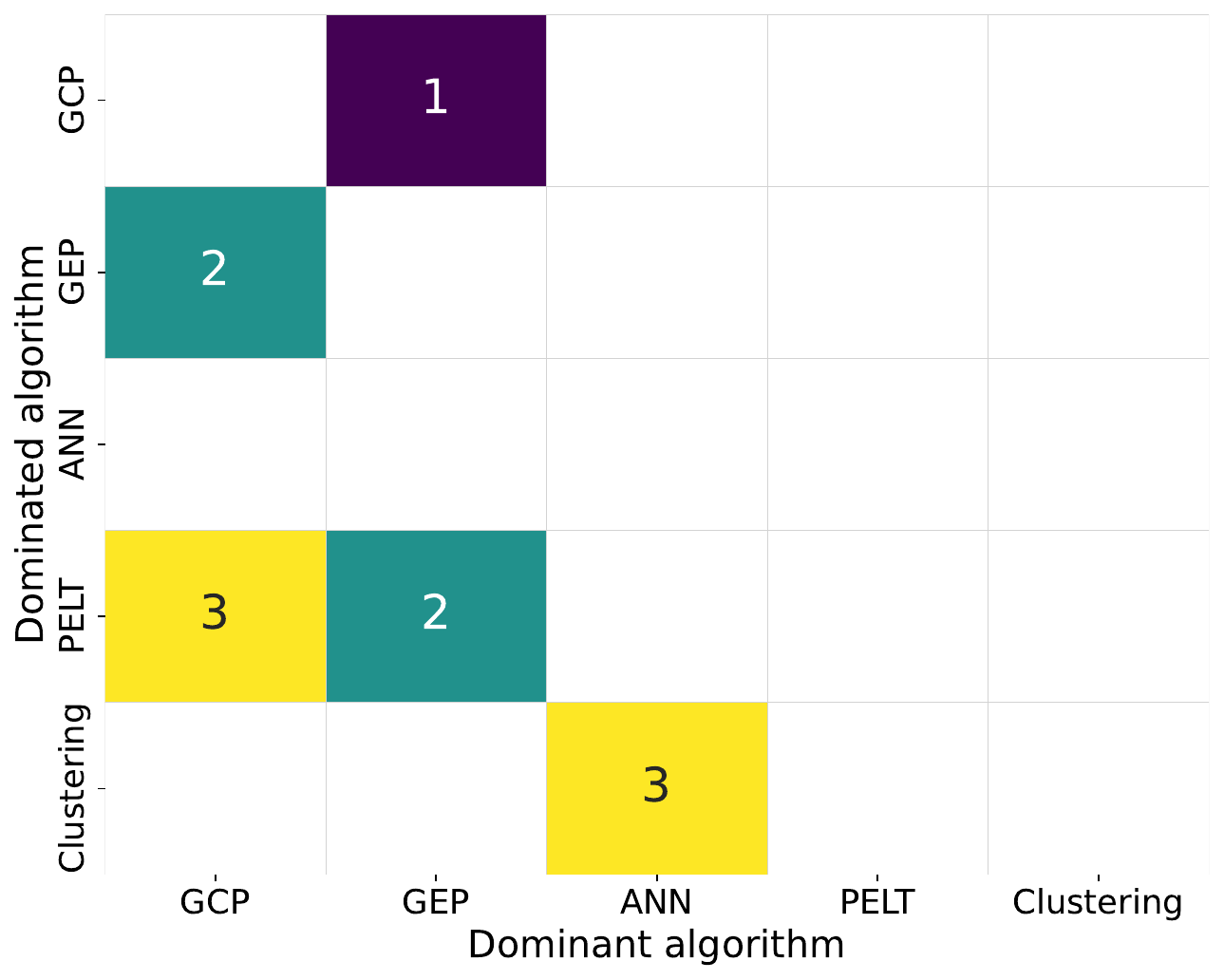}
  \caption{Visualization of pairwise comparisons of \ac{CPD} algorithms. The x-axis reports the dominant algorithms, while the y-axis reports the dominated algorithms. Each cell reports the number of times that a dominant algorithm is found to outperform a dominated algorithm in the surveyed literature}
  \label{fig:cpd heatmap}
\end{figure}

Figure \ref{fig:cpd heatmap} shows the algorithms used in the surveyed \ac{CPD} works, reporting, in each cell, the number of times that a dominant algorithm is found to outperform another approach. The algorithms are sorted in decreasing order of dominance. GEP denotes \ac{GBM}-Euclidean-\ac{PELT} and GCP indicates \ac{GBM}-\ac{CORT}-\ac{PELT}.

\ac{GBM} is a \ac{ML} model used in \cite{Touzani2019}, where it shows its effectiveness for pre-processing.
The \ac{ANN} also looks promising, as it achieves higher performance with respect to a clustering-based method \cite{van2018detection}. Another algorithm is \ac{CPA} \cite{Pereira2018}, which is not compared with the other methods. Also, \ac{GBM} is not compared to \acp{ANN} nor to any clustering-based method. As evident from Figure \ref{fig:cpd heatmap}, further research is needed to determine alternative approaches' relative efficacy.

{\subsection{Time series characteristics}}

{In the reviewed research, a limited number of studies use public  datasets, which hinders a direct examination of the characteristics of the time series. However, a commonality in features and behaviours across diverse datasets is notable. Key features frequently analyzed in machines such as refrigerators, heat pumps or chillers include condenser temperature, evaporator temperature and compressor motor current (particularly in \ac{FD} and \ac{FP}). Energy consumption data, instead, is common in forecasting. The correlation of  the characteristics of time series to the suitability of the inference methods would require systematic comparative studies, which are absent. Most surveyed works employ simple \ac{ML} algorithms. Even though \ac{DL} leads to better results when compared with simpler \ac{ML} algorithms, most works show that it is possible to achieve good performances (e.g., in terms of accuracy) without resorting to more complex and time-consuming \ac{DL} algorithms. The effectiveness of basic \ac{ML} can depend on the low complexity of time series, as shown in \cite{arxivpred}. Low-complexity time series may also  make the use of rule-based approaches feasible \cite{Wu2021, petri2023anomalearn} and preferable over data driven methods.  However, further studies are needed to evaluate time series complexity and to correlate it to the effectiveness of diverse   approaches for the surveyed tasks.}

\section{Issues and Research Directions}
\label{sec:issues_research_directions}

\subsection{Open issues}

\begin{table}[]
\centering
\caption{The main open issues found in this survey, divided by task}
\label{tab:open_issues}
\begin{tabular}{@{}lcccc@{}}
\toprule
\textbf{Issue} & \textbf{\ac{FD}} & \textbf{\ac{FP}} & \textbf{Forecasting} & \textbf{\ac{CPD}} \\ \midrule
\textbf{Public dataset availability} & \xmark & \xmark & \xmark & \xmark \\
\textbf{Benchmarking} & \xmark & \xmark & \xmark & \xmark \\
\textbf{Scarcity of real data} & \xmark & \xmark & \xmark & \xmark \\
\textbf{Class imbalance} & \xmark & \xmark & & \xmark \\
\textbf{Generalization} & \xmark & \xmark & \xmark & \\
\textbf{Interpretability} & \xmark & \xmark & \xmark & \xmark \\
\textbf{Scarcity of comparisons} & \xmark & \xmark & \xmark & \xmark\\ \bottomrule
\end{tabular}
\end{table}

Table \ref{tab:open_issues} summarizes the open issues found in this survey, divided by task. Most of the tasks share the same open issues.

\begin{itemize}
  \item \textbf{Public dataset availability}: As shown in Tables \ref{tab:fault_detection_results}, \ref{tab:fault prediction results}, \ref{tab:forecasting_datas_and_results} and \ref{tab:change point detection data and results} there are few publicly available datasets. The main reason may be industrial non-disclosure policies, defined to prevent the disclosure of real or simulated data revealing details of proprietary industrial implementations.
  
  \item \textbf{Benchmarking}: The lack of widely adopted datasets prevents the comparison of different approaches and the reliable evaluation of the relative performances of alternative methods. A benchmark should comprise a large and representative set of samples, addressing diverse types of faults, change points, or future evolution of the system. As noted in \cite{Wu2021}, in the time series analysis field, some datasets contain anomalies that can be detected trivially, which may hinder the effective evaluation of the research progress. A benchmark should contain both simple and more complex patterns. Inspiration can be taken from existing benchmark datasets in the general domain of time series analysis, such as, for example, the UCR Time Series Anomaly Archive \cite{Wu2021}, which contains multiple time series with varying degrees of complexity designed to challenge the analysis methods, and the MetroPT dataset, which is published with the intent of fostering the benchmarking of predictive maintenance models in the transportation sector  \cite{https://doi.org/10.48550/arxiv.2207.05466}.
  
   \item \textbf{Scarcity of real data}: Especially in the case of \ac{FD} and \ac{FP}, the rarity of failures makes the collection of a large amount of real data unfeasible and most works either use completely simulated datasets or inject synthetic failures in real data. Models trained with simulated data for compressor-based machines may not exhibit the same level of performance when applied to real-world data, as shown for the \ac{FD} task in \cite{Soltani2022}.
   
  \item \textbf{Class imbalance}: \ac{FP}, \ac{FD} and \ac{CPD} are subject to class imbalance, caused by the abundance of normal data compared to anomalies and faults \cite{Seol2023, WANG2013}. Several techniques address the class imbalance problem. They include dataset balancing through oversampling \cite{Gosain2017} and undersampling \cite{XuYingLiu2009}, and the use of evaluation metrics not affected by the imbalance \cite{Ahsan2021}. However, class imbalance is still an open issue because very large datasets are needed to obtain a set of representative anomalous behaviours.
  
  \item \textbf{Generalization}: Dealing with real-world datasets presents challenges, including the difficulty in obtaining time series with faults occurring across a wide range of operating conditions, limiting the diversity of available data \cite{Yan2020}. The lack of intra-dataset diversity leads to the development of ad-hoc models. Only a few works use \ac{TL} to adapt existing models to different scenarios \cite{Qian2020, Zhu2021}. As a result, the question of whether a model developed and tested on a specific dataset retains its capability when applied to a different instance of the same problem (e.g., a compressor-based machine of the same class but of a different producer) remains largely unanswered.

  \item \textbf{Interpretability}: Most surveyed works treat compressor-based machines as black boxes, which makes the interpretability of predictions particularly challenging because the outcome of the predictive models cannot be related to the physical properties of the appliance. Some works use simple \ac{ML} algorithms, which are more interpretable but, in the surveyed works, less effective than \ac{DL} algorithms, as shown in Figures \ref{fig:fd heatmap}, \ref{fig:fp heatmap} and \ref{fig:forecasting heatmap}. The trade-off between the inference's quality and the outcomes' interpretability is still an open challenge. 

  \item \textbf{Scarcity of comparisons}: Most works compare the advocated techniques with only a few previous methods, as shown in Section \ref{sec:Tasks and methods}, which makes it impossible to evaluate the effective contribution brought by a novel approach to the solution of a task.
\end{itemize}

\subsection{Emerging research directions}

The identified open issues can also be viewed as opportunities for researchers to contribute innovative ideas for addressing the inference tasks related to the analysis of compressor-based machines. In this Section we highlight a number of emerging research directions particularly relevant for the progress in the field.

\begin{itemize}
    \item \textbf{Creation of public datasets with real data}: In the time series anomaly detection field, the work in \cite{Wu2021} has introduced the UCR Time Series Anomaly Archive, which provides the community with a benchmark to compare different approaches on diverse datasets. The work in \cite{https://doi.org/10.48550/arxiv.2207.05466} proposes a benchmark dataset for predictive maintenance using real railway data. Finally, the work in \cite{Diallo2021} defines the requirements of a benchmark for failure prediction and describes existing publicly available benchmarks (e.g., \cite{Teng2016, Yoo2018}). A similar effort would be extremely valuable in the field of compressor-based machines, as the lack of public datasets with real data is one of the main open issues in all the tasks analyzed in this survey. Developing and extending tools to compare a large set of techniques would also be beneficial, similar to ODIN, presented in \cite{Torres2022, Zangrando2022, Zangrando2023}.

    \item \textbf{Generalization}: Foundation Models are large-scale general-purpose neural networks pre-trained in an unsupervised manner on massive datasets \cite{rasul2023lag, Yeh2023}. They have shown their effectiveness in various tasks, such as text-to-image generation, news classification and cross-modal retrieval \cite{Fei2022}. Still, Foundation Models' reasoning capabilities need to be assessed when applied to specific tasks (e.g., clinical tasks \cite{Wornow2023}), making interpretability a crucial part of the evaluation \cite{https://doi.org/10.48550/arxiv.2108.07258}. Techniques such as \ac{TL} can also be beneficial, as models developed for a specific problem can be reused as the starting point for addressing similar problems \cite{Sarmas2022, TapiaRivas2024}, but only a few of the surveyed works consider this approach (e.g., \cite{Zhu2021, Qian2020}). Ensemble Learning is a \ac{ML} paradigm where multiple simpler learners (base learners) are trained to solve the same problem and their results are finally combined. The generalization ability of an ensemble is usually stronger than that of the base learners \cite{Zhou2009}. They have shown their effectiveness in diverse fields, including hydrology \cite{Nguyen2022} and medicine \cite{Saito2023}. The work in \cite{Hao2024} also shows their effectiveness in long-term multivariate time series forecasting, which is crucial for the compressor-based machines case.
    
    \item \textbf{Intrepretability}: In the field of predictive maintenance, the work in \cite{Vollert2021} identifies the challenges of interpretable \ac{ML} for predictive maintenance, and highlights the connection between interpretability and trust in models in industrial settings. It notes that the vast majority of interpretable \ac{ML} research is conducted on images, while interpretable \ac{ML} for time series is the subject of future research. Compressor-based machines show a deterministic, yet complex, behaviour, as other physical systems. For modelling physical phenomena, \acp{PINN} have shown their effectiveness in several fields, including climate \cite{Chen20232} and electronics \cite{Wang2023}. \ac{SR} is another approach that has shown its effectiveness in physics \cite{PhysRevE.94.012214}, as it can devise analytical treatable models of the results. Finally, \acp{GNN} can help considering the correlation between variables \cite{Xiao2023}. Combining such approaches would make models inherently interpretable and thus more reliable.
    \item {\textbf{Analysis of multi-modal data}: The diffusion of 5G connectivity in both industrial and domestic environments enables scenarios in which  monitoring an appliance goes beyond acquiring   the values of status variables over time. The benefits  of employing multi-modal data for detection and prediction tasks have been demonstrated in the most disparate  fields \cite{Fernandez2024, Boehm2022, PinciroliVago2023, Fan2022} with various works specifically investigating the advantages of combining numerical time series and images \cite{PinciroliVago2023, Fan2022, Morgan2022, Morgan2023}.
    In the field of appliance monitoring, the work in \cite{Feng2023vibrat} shows that surface degradation in gears can be observed in images and in time series. In compressor-based machines,  noise analysis  can be exploited  to  determine the working status of the compressor \cite{N2021ModelingAF,Verma2016,1529092}. Also olfactory data acquisition, enabled by artificial nose sensors \cite{Kim2022}, can help detect malfunctions associated with the emission of fumes \cite{Gui2024} and  the insurgence of overheating and of burning materials in compressor-based machines \cite{Beasley2018, sengupta2014modelling}. Studying the effectiveness and the cost-benefit trade-offs of different combinations of multi-modal data for detection and prediction tasks  constitutes a relevant and promising novel research direction.}

\item  {\textbf{Application of signal processing  to gear fault diagnosis}: gears  are fundamental components of dynamic compressors and are often involved in machine failures. This makes gear health status and \ac{RUL} estimation relevant for compressor-based machine monitoring. Signal processing techniques are an emerging area of application to gear fault diagnosis and can be used in conjunction with other data-driven methods to improve the prediction and characterization of failures in compressor-based machines.  The work in \cite{Feng2023} demonstrates that the \ac{RUL} of gears can be estimated using \ac{TE} (i.e., the relative angular difference between the rotations of two gears) using a \ac{ML} algorithm. The work in \cite{Feng2022} diagnoses faults in a wind turbine planetary gearbox and demonstrates that a novel approach based on Vold-Kalman filtration \cite{Vold1995, 202548b1-0892-3155-8ebe-0a20c956b55e} can extract and track non-stationary vibration harmonics of rotating machines. The work in \cite{Feng2023cyclo} considers cyclostationary processes (i.e., signals with statistical characteristics that vary periodically \cite{Gardner2006}) to evaluate gear wear propagation, represent the surface wear distribution and assess the surface wear severity. To accomplish this task, it uses raw vibration signals, formed by a deterministic and a random component, and uses an adaptive filter to extract the deterministic component. The study of gear surface wear progression is considered also in \cite{Feng2023vibrat}, which also uses vibration data, proposes a novel \ac{HI} and a \ac{GRU} for \ac{RUL} prediction.}

\end{itemize}

\section{Conclusions}
\label{sec:conclusions}
This survey addresses \ac{FD}, \ac{FP}, forecasting and \ac{CPD} for compressor-based machines. For each task, algorithms belonging to different families have been compared. \ac{FD} is essential for preventing machine downtime and more serious failures. For such a task, \ac{DL} algorithms are commonly used, with techniques such as \ac{SVM}, \ac{1D-CNN} and \ac{LSTM} networks proving the most effective. \ac{SVM} is noted for its non-linear classification abilities, \ac{LSTM} for capturing time-dependent patterns and \ac{1D-CNN} for identifying local dependencies. However, real-world dataset availability and the scarcity of comparisons between top-performing algorithms are still open challenges.
\ac{FP} is crucial in predicting faults in advance. It uses methods such as autoencoders, XGBoost, \ac{RF}, \ac{SVM} and \ac{KNN}. The main issues are the absence of public datasets and the need for comparison with existing approaches.
Forecasting exploits historical and current data to predict the future value of the variables of interest. \ac{SVM}, \ac{LSTM} and \ac{GRU} have been employed more frequently, with \ac{LSTM} and \ac{GRU} delivering superior results due to their ability to handle long-term data dependencies. The scarcity of public datasets and the absence of comparisons with effective approaches are open issues. \ac{CPD} identifies status changes in machines, using such methods as \ac{GBM} combined with \ac{CORT} dissimilarity measures, clustering and \acp{ANN}. Challenges include limited datasets with real data and a lack of comparisons with existing research.
After presenting and evaluating the status of the art for all the tasks, we distil the main research issues that are still open and hint at some promising research directions that stem from the identified research gaps. 
To conclude, the \ac{FD}, \ac{FP}, forecasting and \ac{CPD} tasks are crucial for managing industrial and residential compressor-based machines in various domains, but the surveyed works do not always compare the proposed or tested methods with other state-of-the-art approaches. Moreover, since most datasets are proprietary and not published and community-shared benchmarks do not exist, comparing different approaches fairly is still out of reach.

{Based on our analysis, several directions for future works emerge.
The development of practical approaches for \ac{FD}, \ac{FP}, Forecasting, and \ac{CPD} requires a specific focus on model generalization and interpretability, which involves exploring how \ac{ML} and \ac{DL} techniques adapt to changing machine behaviours over time. Secondly, there is an urgent need for more extensive and diverse public datasets and agreed-upon benchmarks to enable fair comparison of approaches and consolidate the progress in the field. Lastly, there is significant potential in the creation of innovative hybrid models that combine multi-modal data sources, complementary  \ac{ML},  \ac{DL} and signal processing techniques, in order to improve the performance and applicability of the data-driven methods for compressor-based machine detection and prediction tasks.}

\section*{Statements and Declarations}

\paragraph{Conflict of interest}
The authors declare that there is no conflict of interest.

\paragraph{Author Contributions}
Conceptualization, F.F., N.O.P.V. and P.F.; methodology, F.F., N.O.P.V. and P.F.; formal analysis, F.F., N.O.P.V. and P.F.; investigation, F.F., N.O.P.V. and P.F.; writing—original
draft preparation, F.F., N.O.P.V. and P.F.; writing—review and editing, F.F., N.O.P.V. and P.F. All
authors have read and agreed to the published version of the manuscript.

\paragraph{Funding}
This work has been supported by the European Union's Horizon 2020 project PRECEPT, under grant agreement No. 958284.

\bibliography{sn-bibliography}
\end{document}

%% file: acronym.tex
\begin{acronym}
\acro{1D-CNN}{One Dimensional Convolutional Neural Network}
\acro{AARE}{Average Absolute Relative Error}
\acro{AHU}{Air Handling Unit}
\acro{A-LSTM}{Attention-LSTM}
\acro{ANFIS}{Adaptive Network-based Fuzzy
Inference System}
\acro{ANN}{Artificial Neural Network}
\acro{ASHRAE}{American Society of Heating, Refrigerating and Air-Conditioning Engineers}
\acro{AutoML}{Auto Machine Learning}
\acro{Bi-LSTM}{Bidirectional Long Short Term Memory}
\acro{BN}{Bayesian Network}
\acro{BOA}{Bayesian Optimization Algorithm}
\acro{BPNN}{Back Propagation Neural Network}
\acro{CNN}{Convolutional Neural Network}
\acro{COP}{Coefficient of Performance}
\acro{CORT}{Correlation-based Overlapping Rotation Transformation}
\acro{CPA}{Change Point Analysis}
\acro{CPD}{Change Point Detection}
\acro{CV(RMSE)}{Coefficient of Variation of the Root Mean Square Error}
\acro{CWGAN}{Conditional Wasserstein Generative Adversarial Network}
\acro{DANN}{Domain Adversarial Neural Networks}
\acro{DJ}{Decision Jungle}
\acro{DL}{Deep Learning}
\acro{DNN}{Deep Neural Network}
\acro{DRNN}{Deep Recurrent Neural Network}
\acro{DT}{Decision Tree}
\acro{DTO-DRNN}{Deep Transition Output - Deep Recurrent Neural Network}
\acro{DTWD}{Dynamic Time Warp Distance}
\acro{EKF}{Extended Kalman Filter}
\acro{ELM}{Extreme Learning Machine}
\acro{EMD}{Empirical Mode Decomposition}
\acro{FCDNN}{Fully Connected Deep Neural Network}
\acro{FD}{Fault Detection}
\acro{FDD}{Fault Detection and Diagnosis}
\acro{FF-AE}{Feed Forward Autoencoder}
\acro{FFNN}{Feed Forward Neural Network}
\acro{FI}{factory-installed}
\acro{FITNET}{Fitting Neural Networks}
\acro{FP}{False Positive}
\acro{FP}{Fault Prediction}
\acro{GA}{Genetic Algorithm}
\acro{GAN}{Generative Adversarial Network}
\acro{GB}{Gradient Boosting}
\acro{GBM}{Gradient Boosting Machine}
\acro{GNN}{Graph Neural Network}
\acro{GP}{Genetic Programming}
\acro{GPR}{Gaussian Process Regression}
\acro{GRU}{Gated Recurrent Unit}
\acro{GSHPS}{Ground Source Heat Pump System}
\acro{GT}{Ground Truth}
\acro{HI}{Health Indicator}
\acro{HGSHP}{Hybrid Ground Source Heat Pump}
\acro{HTC}{Heat Transfer Coefficient}
\acro{HVAC}{Heat, Ventilation, and Air Conditioning}
\acro{IoT}{Internet of Things}
\acro{ISSA}{Improved Sparrow Search Algorithm}
\acro{KNN}{K-Nearest Neighbors}
\acro{LDA}{Linear Discriminant Analysis}
\acro{LMB-NN}{Levenberg-Marquardt Backpropagation Neural Network}
\acro{LR}{Linear Regression}
\acro{LSFM}{Least Square Fitting Method}
\acro{LSTM}{Long Short-Term Memory}
\acro{LSTM-AE}{LSTM-AutoEncoder}
\acro{MAE}{Mean Absolute Error}
\acro{MAPE}{Mean Absolute Percent Error}
\acro{MAX}{Maximum error}
\acro{MBE}{Mean Bias Error}
\acro{MC-SVM}{Multi-Class Support Vector Machine}
\acro{ML}{Machine Learning}
\acro{MLP}{Multilayer Perceptron}
\acro{MLR}{Multi-linear Regression}
\acro{MPMR}{Mini-Max Probability Machine Regression}
\acro{mRMR}{min-Redundancy Max-Redundancy}
\acro{MSE}{Means Square Error}
\acro{NARX-NN}{Nonlinear Autoregressive Exogenous Neural Network}
\acro{NB}{Naïve Bayes}
\acro{NRE}{Non-Regular Event}
\acro{OSVM}{One-Class Support Vector Machine}
\acro{OVA}{One-Versus-All multiclass classification}
\acro{PCA}{Principal Component Analysis}
\acro{PELT}{Pruned Exact Linear Time}
\acro{PINN}{Physics-Informed Neural Network}
\acro{PSO}{Particle Swarm Optimization}
\acro{PW}{Prediction Window}
\acro{RBF}{Radial Basis Function}
\acro{RF}{Random Forest}
\acro{RFECV}{Recursive Feature Elimination based on Cross Validation}
\acro{RMSE}{Root Mean Square Error}
\acro{RNN}{Recurrent Neural Network}
\acro{ROSVM}{Recursive One-class Support Vector Machine}
\acro{RW}{Reading Window}
\acro{RUL}{Remaining Useful Life}
\acro{S,DTO-DRNN}{Stacked, Deep Transition Output - Deep Recurrent Neural Network}
\acro{SARIMAX}{Seasonal Auto-Regressive Integrated Moving Average with eXogenous factors}
\acro{S-DRNN}{Stacked Deep Recurrent Neural Network}
\acro{Seq2Seq LSTM}{Sequence-to-Sequence Long Short Term Memory}
\acro{SL}{Severity Level}
\acro{SNN}{Shallow Neural Network}
\acro{SR}{Symbolic Regression}
\acro{SVM}{Support Vector Machine}
\acro{SVR}{Support Vector Regression}
\acro{TE}{Transmission Error}
\acro{TIME}{Evaluation time}
\acro{TL}{Transfer Learning}
\acro{TU}{Terminal Unit}
\acro{VAE}{Variational AutoEncoder}
\acro{VMD}{Variational Mode Decomposition}
\acro{WM}{Wang and Mendel's fuzzy rule learning method}
\end{acronym}